\documentclass{article}

\usepackage{arxiv}

\usepackage{microtype}
\usepackage{graphicx}
\usepackage{caption}
\usepackage{subcaption}
\usepackage{booktabs} 
\usepackage{amsmath}
\usepackage{bbm}

\usepackage{enumitem}

\usepackage[utf8]{inputenc} 
\usepackage[T1]{fontenc}    
\usepackage{hyperref}       
\usepackage{cleveref}
\usepackage{url}            
\usepackage{booktabs}       
\usepackage{amsfonts}       
\usepackage{nicefrac}       
\usepackage{microtype}      

\usepackage[noend]{algpseudocode}
\usepackage{algorithm}
\usepackage{algorithmicx}

\usepackage{soul}
\usepackage{multirow}

\title{Toward A Neuro-inspired Creative Decoder}

\newcommand*\samethanks[1][\value{footnote}]{\footnotemark[#1]}

\author{
  Payel Das \thanks{Payel Das and Brian Quanz contributed equally to this work and are corresponding authors.  } \\
  IBM Research, Yorktown Heights, NY, USA\\
  Department of Applied Physics and Applied Math, Columbia University, New York, NY, USA\\
  \texttt{daspa@us.ibm.com} \\
  \And
  Brian Quanz \samethanks[1] \\
  IBM Research\\
  Yorktown Heights, NY, USA \\
  \texttt{blquanz@us.ibm.com} \\
  \AND
  Pin-Yu Chen \\
  IBM Research\\
  Yorktown Heights, NY, USA \\
  \texttt{Pin-Yu.Chen@ibm.com} \\
  \And
  Jae-wook Ahn \\
  IBM Research\\
  Yorktown Heights, NY, USA \\
  \texttt{jaewook.ahn@us.ibm.com} \\
  \And
  Dhruv Shah\\
  IBM Research\\
  Yorktown Heights, NY, USA \\
  \texttt{Dhruv.Shah@ibm.com} 
}

\begin{document}

\maketitle

\begin{abstract}
Creativity, a process that generates novel and meaningful ideas, involves increased association between task-positive (control) and task-negative (default) networks in the human brain. Inspired by this seminal finding, in this study we propose a creative decoder within a deep generative framework, which involves direct modulation of the neuronal activation pattern after sampling from the learned latent space. The proposed approach is fully unsupervised and can be used off-the-shelf. Several novelty metrics and human evaluation were used to evaluate the creative capacity of the deep decoder. Our experiments on different image datasets (MNIST, FMNIST, MNIST+FMNIST, WikiArt and CelebA) reveal that atypical co-activation of highly activated and weakly activated neurons in a deep decoder promotes generation of novel and meaningful artifacts.  
\end{abstract}


\section{Introduction}
\label{Introduction}

Creativity is defined as a process that produces novel and valuable (\textit{aka} meaningful) ideas \cite{boden2004creative}.
In early days of computational creativity research, expert feedback  \cite{graf1995interactive} 
or evolutionary algorithms with hand-crafted fitness functions \cite{dipaola2009incorporating}
were used to guide a model's search process to make it creative. However, those methods reportedly lack exploration capability. 

Data-driven approaches like deep learning open a new direction -- enabling the study of creativity from a knowledge acquisition perspective. Deep Dream \cite{szegedy2015going} and Deep Style Transfer \cite{gatys2015neural} have aroused substantial interest to employ deep learning-based methods in computational creativity research. Only recently, novelty generation using powerful deep generative models, such as Variational Autoencoders (VAEs)\cite{kingma2013auto,rezende2015variational}  and Generative Adversarial Networks (GANs) \cite{goodfellow2014generative}, have  been attempted. These are designed to build a model from and generate known objects (\textit{i.e.}, images).
However, such models discourage out-of-distribution generation to avoid instability and minimize spurious sample generation, limiting their potential in creativity research. In fact,  \cite{kegl2018spurious}  shows that getting rid of ``spurious'' samples completely can limit the generative modeling capacity. Therefore, new approaches to enhance the creative capacity of generative models are needed.   

\begin{figure*}[t] 
\centering
\includegraphics[width=6.2 in]{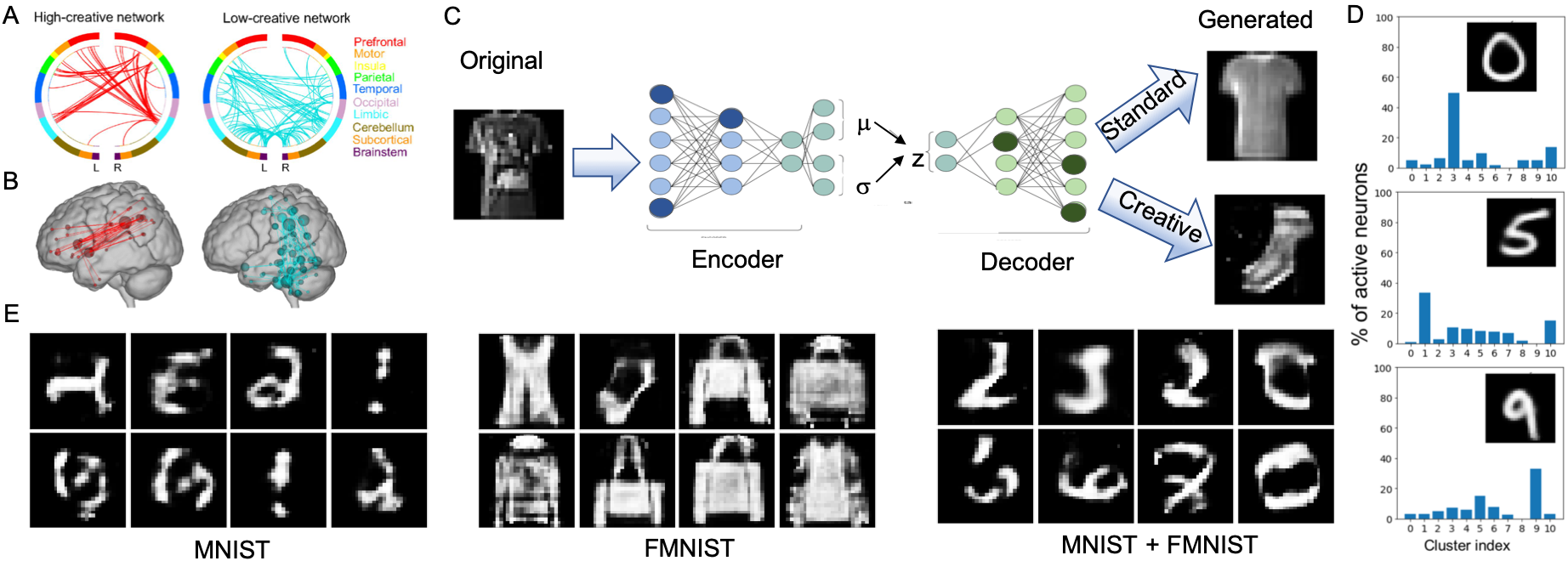}  
\caption{\textbf{A-B.} Depictions (A: circle plots, B: glass brains) of high- and low-creative networks in human brains with their highest degree nodes. Circle plot colors correspond to brain lobes: L, left hemisphere; R, right hemisphere. Adapted from \cite{beaty2018robust}. \textbf{C.} Depiction of a VAE model with our neuro-inspired \textit{creative} decoder. Normally, a small fraction of neurons in each hidden layer are low-active (dark color). Inspired by neural basis of creativity, we activate those ``low-active'' (task-negative) neurons to induce coupling between task-positive and task-negative neurons  during ``creative'' decoding.  \textbf{D.} Class activation of neuronal clusters in an MNIST VAE, used by the cluster-level activation version of our method that models the neuronal networks with activation clusters and works by co-activating ``low-active'' neuronal activation cluster(s) (representing the task-negative network), suggesting cluster activation patterns are associated with high-level concepts (e.g., digit classes / components).  \textbf{E.} Samples generated by the proposed \textit{creative} decoder that were human-annotated as creative with high confidence. }
\label{Creativity}
\end{figure*}

One path is to get inspiration from the cognitive processes associated with human creativity.  
How does the human brain produce creative ideas? It is an interesting question and a central topic in cognitive neuroscience research. A  number of recent neuroimaging studies \cite{beaty2018robust,shi2018large,gao2017exploring} suggest stronger coupling of default mode network and executive control network in creative brains across a range of creative tasks and domains, from divergent thinking to poetry composition to musical improvisation (see Fig. \ref{Creativity}A-B). Brain networks are the large-scale communities of interacting brain regions, as revealed by the resting-state functional correlation pattern; 
 these networks correspond to distinct functional systems of the brain.  Default (task-negative) mode network
is associated with spontaneous and self-generated thought, and, therefore implicated in idea generation. 
Control (task-positive) network, in contrast, is associated with cognitive processes requiring externally directed attention. 

Default and control networks often exhibit an antagonistic relation during rest and many cognitive tasks, including working memory \cite{anticevic2012role}.
 This antagonistic relation likely reflects suppression of task-unrelated thoughts during cognitive control. Dynamic coupling of default and control networks has been previously reported during goal-directed, self-generated thought processes
 \cite{spreng2015autobiographical}. 
 Recently \cite{beaty2016creative} proposed that stronger coordination between default and control networks contributes to creative idea generation. 

\textbf{Research Question:}
Motivated by neuroimaging findings suggesting stronger coupling between task-positive and task-negative neurons in creative brains, this work attempts to induce creativity in deep generative models by proposing a \textit{creative} decoder.   In a nutshell,  the   \textit{creative} decoder aims to generate creative samples from the original latent (concept) space by favoring atypical co-activation of high- and low-active neurons (neuron groups derived by roughly modeling the task-negative and task-positive concepts), while the generative model training remains unchanged.

It is widely accepted that creativity is a combination of  novelty and value. However, value determination of a novel artefact is  non-trivial and design of clear-cut creativity evaluation schemes is believed to be as important as developing \textit{creative} generative methods \cite{cherti2017out}. 
As the goal is to generate meaningful novelty, not trivial noise, traditional metrics based on likelihood/distance are of limited use. Nevertheless, novelty alone has been found to be a better predictor of creativity than value \cite{diedrich2015creative}, and consistently,  stringent evaluation is found to hinder creativity in  brain \cite{beaty2017creative}.

In the present study, we employ human annotation for creativity evaluation, in addition to using a number of surrogate metrics (supervised and unsupervised) for  novelty estimation, of the samples generated by the \textit{creative} decoder.  A VAE model was used as the base generative framework (see Fig \ref{Creativity}C).  
We show the performance of  the proposed method against MNIST digits
, FMINST fashion objects 
, and on a combined  MNIST plus FMINST dataset. We also present results on the WikiArt art images and CelebA faces. 

Our main contributions are:

\noindent$\bullet$ Inspired by neuroimaging findings, we propose a VAE model  with a \textit{creative} decoder, which generates novel and meaningful samples from the learned representation space by employing an atypical activation pattern during decoding. 
This modified generation scheme \textit{does not require any data labels} and can be adapted to any decoder/generator.
    
\noindent$\bullet$ Different schemes (\textit{correlation-based}, \textit{cluster-based}, and \textit{low-active})  of stochastic co-activation of ``on'' and ``off''  neurons during decoding are introduced and tested in terms of  creative capacity and compared against a number of unsupervised baseline  decoding methods. Results show that enhanced creativity can result from the neuro-inspired atypical activation as opposed to a simple random or structured noise effect.  

\noindent$\bullet$ The creativity  of generated samples was evaluated using human annotation as well as several surrogate metrics for novelty detection (\textit{e.g.},  in-domain classifier score and reconstruction distance). Our analyses suggest that a combination of existing surrogate metrics largely captures the notion of creativity.  
    
\noindent$\bullet$ A key advantage of our method is its immediate applicability  to any off-the-shelf decoder/generator model, as it requires no retraining,  access to abnormal samples, or additional categorical information for creative generation.   To our knowledge, this is the first work that aims to enhance creative capacity of a deep generative model in a neuro-inspired manner\footnote{The codes and data will be made available upon publication.}.

\section{Related Work}
\textbf{Deep Generative Models For Novelty Generation} 

\cite{nguyen2015innovation} proposed the deep neural network (DNN)-based \textit{innovation engine}  algorithm,  demonstrating that evolving toward many objectives simultaneously approximates divergent search and encourages novel generation. It is composed of: (1) a diversity-promoting evolutionary  algorithm (EA) that generates  novel instances based on pre-set features, and (2) a deep   classifier to determine if generated instances are interesting and should be retained. 
The main differences between \textit{innovation engine} and \textit{creative} decoder are use of a data-driven generative model instead of EA  and inducetion of novelty without  a set of pre-set features. 

Another line of work is few/one-shot learning and generation \cite{lake2015human,rezende2016one,clouatre2019figr}, a step toward out-of-class generation. \cite{lake2015human} trains a model with Bayesian Program Learning, which represents concepts (pen strokes) as simple probabilistic programs and hierarchically combines them to generate images (letters). They showed that the model can be trained on a single image of a previously unseen class and generate novel samples of it.  \cite{rezende2016one}  developed a class of sequential generative models to achieve one-shot generation. 
They showed that a combination of sequential generation and inference in a VAE model 
and attention \cite{bahdanau2014neural}
during inference enables generation of compelling alternative variations of images after seeing them once. Recently,  \cite{clouatre2019figr} combined GANs 
with meta-learning \cite{nichol2018first} to find GAN parameters that can generate a random instance with little data and training. The difference with this line of work is that ours does not need any novel exemplar during test time for novel generation. 

Next, the creative adversarial network (CAN), a GAN variant, attempts to generate novel artistic images by minimizing deviation from an art distribution while maximizing style ambiguity \cite{elgammal2017can}. CAN is based on a psychology concept that exploits style ambiguity as a means to increase arousal potential. 
Unlike our proposed model, CAN needs data-label pairs as input to generate novel samples. 

\cite{cherti2016digits} generated new images that are not digit-like by performing operations like crossover and mutation, in the latent space of MNIST digits learnt using an autoencoder, or by iteratively refining a random input such that it can be easily reconstructed by the trained autoencoder. Their method needs  human feedback to make the network creative-explorative, which is different from our method, as \textit{creative} decoder exploits a reported neural basis of creativity.

\section{Evaluation of Creativity: Machine and Human Perception}
\textbf{Human Evaluation of Creativity}

 The ultimate test of creativity is through inspection by humans.  Additionally, human labeling has been used to evaluate deep generative models    \cite
 {dosovitskiy2016learning,lopez2018human} or as a part of the generative pipeline \cite{lake2015human,salimans2016improved}. Although human judgment of creativity suffers from several drawbacks (it is costly, cumbersome, and is subjective), it is still crucial to check how humans perceive and judge generated images. We used an in-house  annotation tool to evaluate creativity of the generated samples. Given an image, the tool had four options to choose from - `not novel or creative', `novel but not creative', `creative', and `inconclusive'. 
 Annotators did not have access to the knowledge of the decoding scheme at the time of annotation, but were primed on the training dataset. 
 
\textbf{Surrogate Metrics of Novelty.}
Novelty detection techniques can be broadly categorized in five  categories: (i) probabilistic, (ii) distance-based, (iii) reconstruction-based, (iv) domain-based, and (v) information-theoretic techniques. 
 Below, we outline the novelty  metric families used for evaluation. 

\mathchardef\mhyphen="2D

\textbf{Reconstruction distance.} 
Reconstruction distance based on encoder-decoder architectures has been leveraged for novelty detection 
\cite{wang2018generative}. For image $x$ and corresponding latent (encoded) vector $z$,  novelty can be estimated from the distance  between $x$ and the closest sample the VAE can produce from $z$. Therefore,  
$D_{r}= \min\limits_{z} ||x - E[_{\theta}(x|z)]||_{2}$. Since a trained VAE has a narrow bottleneck, the reconstruction distance of any novel image will be large. 

\textbf{k-Novelty score.}
We compute the $k$th-nearest-neighbor distance between a generated sample and the training dataset in the latent $z$ space and in the input space.
  The kNN novelty score ($kNS$) \cite{ding2014experimental} of a given data point is the normalized distance to its $k$th nearest neighbor (denoted as $kNN(\cdot)$), which is defined as following:
   $kNS = \frac{d(x,kNN(x))}{d(kNN(x),kNN(kNN(x)))}.$
For $k=1$, $d$ is an $l_2$ distance. For $k > 1$, $d$ accounts for the expected distance to the $k$th nearest neighbor. We compute $kNS$ for $k$=1 and 5.

\textbf{In-domain Classifier Entropy (ICE) or ``objectness'' .}
Entropy of the probabilities $p$ returned by a trained, multi-class, in-domain classifier has been used to estimate novelty  \cite{hendrycks2016baseline,kliger2018novelty,cherti2017out}, which can be defined as
    $ICE = -\sum_{i}{p_{i}\log_2 p_{i}}.$
A higher value of $ICE$ implies higher novelty. This metric is similar to what  \cite{salimans2016improved} has used to stabilize GAN  training and  \cite{nguyen2015innovation} has employed to encourage novelty search. 

\textbf{In-domain Score (IS) from a one-class classifier.}
One-class classifiers have also been used for novelty detection, as novel classes are often absent during training, poorly sampled or not well defined.  
In this study, for each dataset we train a one-class support vector machine (SVM) classifier 
on the  latent dimensions obtained  from trained VAE model.  Using this one-class classifier, We classify each generated image to get an in-domain score (IS) - the signed difference from the normal-classifying hyperplane.  The lower this value is,  the stronger the outlierness of the image.

\section{Methodologies and Preliminaries}
\label{sec:method}
\textbf{Variational Autoencoder (VAE).} 
We use a VAE \cite{kingma2013auto}
as the base generative model, which trains a model $p(x,z)$ to maximize the marginal likelihood $\log p(x)$ on dataset samples $x$. As the marginal likelihood requires computing an intractable integral over the unobserved latent variable $z$, VAEs introduce an encoder network $q_{\theta}(z|x)$ and optimize a tractable lower bound (the ELBO):
$\log p(x) \geq$  $E[\log p(x|z)] - D_{KL}[q(z)||p(z)]$.
The first term accounts for the reconstruction loss, the second measures KL loss between the encoder's latent code distribution, $p_{\theta}(z)$, and the prior distribution, typically a diagonal-covariance Gaussian.

\textbf{\textit{Creative} decoding scheme.}
To capture the spirit of the atypical neuronal activation pattern observed in a creative human brain, \textit{i.e.}, dynamic interaction between a task-positive (control) and  a task-negative (default) brain network, we propose a probabilistic decoding scheme.  After sampling a $z$,  the proposed method selects decoder neurons based on their activation correlation pattern, and then activates a few ``off" neurons along with the ``on'' neurons.   Three different co-activation schemes were tested, each selecting a set of ``off'' neurons to turn ``on'' using different grouping criteria to represent ``task-negative'': (1) correlation-based -- random selection from the pool of  ``off'' neurons that are most anti-correlated with the firing neurons across the training data; (2) cluster-based (Fig. \ref{Creativity}D) -- ``off'' neurons were chosen at random from the neuronal clusters (obtained by clustering the neuronal activation map) with low percent of cluster active during decoding; and (3) ``low-active''  based -- random selection from ``off'' neurons that were weakly-activated in the trained decoder across most of the training data. 
Another version of the ``low-active'' method only selected  low-active neurons without any significant class specificity (non-specific low-active method).  

In the following, due to the space constraint we only provide a detailed description of the low-active method; other variants (cluster-based, correlation-based and non-specific low-active method) were found to perform in a similar fashion.

\textbf{Preliminaries.}
Let $d_{j}^{k}(z)$ represent the output of the $j$th neuron of the $k$th layer, given input $z \in \mathcal{Z}$ to the decoder $p_\theta$.  
We further use the short-hand, $d_{ji}^k = d_{j}^{k}(z_i)$, where $z_i$ is the encoding of the $i$th training data point.  Then the \emph{percentage activation} of neuron $j$ in layer $k$ is 
$a_{j}^{k} = (1/n)\sum_{i=1}^{n} \mathbbm{1}_{d_{ji}^{k} > \tau}$, for a given activation threshold $\tau$.
For RELU activation functions we set $\tau$ to $1\mathrm{e}{-7}$.  Neurons with $a_{j}^{k} \approx 0$ are classified as \emph{dead} neurons and excluded from neuronal manipulation.  Additionally, for the $i$th input we call neuron $j$ of layer $k$ ``active'' or ``on'' if $d_{ji}^{k} > \tau$, and ``inactive'' or ``off'' otherwise.
Let $\vec{d_{j}^{k}}$ be the vector with $i$th entry $= d_{ji}^{k}$.  Given neuron $j$ and neuron $h$ in layer $k$, let $C^{kjh}=\mathrm{Cov}[\vec{d_{j}^{k}},\vec{d_{h}^{k}}]$ (the covariance matrix). Then we define their correlation $R_{jh}^{k}= C^{kjh}_{01} / \sqrt{C^{kjh}_{00}C^{kjh}_{11}}$  - which are the entries of the layer correlation matrix $R^k$.

\textbf{Neuron flipping.}
We define flipping a neuron ``off'' by setting it to a minimum activation value, \textit{i.e.}, $0$ for RELU activation.  We define the ``on'' value of a neuron $j$ in layer $k$ as $o_{j}^{k} = \lambda \cdot s(\{d_{ji}^{k}| i=1\ldots n\})$, \textit{i.e.}, $\lambda$ denotes a scaling factor for the statistic of training activation values, \textit{e.g.}, $s$ equal to mean, max, \textit{etc}.  Since,  task-negative neurons are  activated at higher levels than usual during creative processes   in human brain, we set $s = \max$ and $\lambda = 2$ in experiments.  

We start with a sample $z \sim p(z)$ and obtain the neuronal activations  for a selected layer  $k$ of the decoder, for which we now use shorthand $d_{jz}^{k} = d_{j}^{k}(z)$.
Additionally let $\mathcal{A}$ be the corresponding set of active or ``on" neurons and $\mathcal{D}$ the set of inactive or ``off" (non-dead) neurons.
During creative decoding, we flip some number or percentage, $\rho$, of a group of ``on'' and/or ``off" neurons in a  layer $k$, either randomly or selectively. Each method modifies $d_{j}^{k}(z)$, and this modified layer output is then passed through the remainder of the decoder, to obtain the final generated values.  

\textbf{``Low-active'' method.} 
The ``low-active'' method (Algorithm \ref{alg:low-act}) imitates the task-negative concept by identifying neurons that typically have low activation across all the training data and turning some number of them on at decode time.  Specifically, a neuron is selected from  the pool of ``off'' neurons that have the lowest percent activations, $a_{j}^{k}$ (defined using a threshold cutoff).
Next, ``low-active'' neurons that are  most  correlated  with the selected neuron are also turned on (randomly selecting the remainder from the low-active group provides similar results).
\begin{algorithm}
\caption{Low-active method}
  \label{alg:low-act}
\textbf{Input:} Layer output $d_{jz}^{k}$; percent activation percentile $\kappa$; fraction of neurons to flip on $\rho$
\begin{algorithmic}
\State $t \gets $ $\kappa$-percentile($\{a_{j}^{k}\}$)
\State $\mathcal{S} \gets \{j | j \in \mathcal{D} \land a_{j}^{k} \leq t\}$
\State $s \gets$ random select from $\mathcal{S}$
\State $\mathcal{S}_{s} \gets {\rm argmax}_{\mathcal{S}'\subset \mathcal{S}, |\mathcal{S}'|=\left \lfloor{\rho|\mathcal{S}|}\right \rfloor}\sum_{h \in \mathcal{S}'}R_{sh}^{k}$
\State $d_{jz}^{k} \gets o_{j}^{k}, \forall j \in \mathcal{S}_s$
\end{algorithmic}
\end{algorithm}

\textbf{Baseline decoding schemes.}
The samples generated by the \textit{creative} decoding schemes were compared with the original training distributions as well as with samples generated by (1) Linear Interpolation in the latent space between training samples that belong to different classes, followed by standard decoding (result not shown), (2) Noisy-decoding: During decoding  a random Gaussian noise was infused in a fraction of neurons, and (3) Random flipping: activation of randomly selected ``off'' neurons (same as our proposed methods but does not apply special criteria to select ``off'' neurons).

\begin{figure*}[t]    
\centering
\includegraphics[width=7 in]{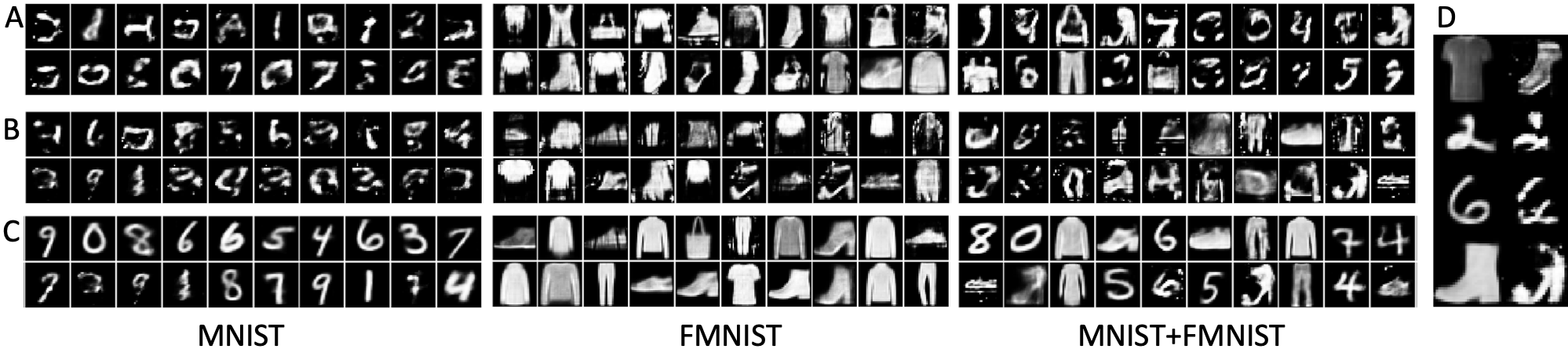}  
\caption{Random VAE-generated samples  human-annotated as (A) creative, (B) novel but not-creative, and (C) not novel or creative. D: How ``creative'' decoding modifies generation (combined dataset): \textbf{left:} regular; \textbf{right:} low-active decoding. 
}
\label{Human}
\end{figure*}

\section{Experimental Details}
\textbf{VAE architecture.}
For F/MNIST the encoder network consisted of 3 fully-connected layers (1000, 500, 250) before the $z$ output (50 for F/MNIST and 100 for the combination), with the decoder architecture the reverse of the encoder. RELU activations were used; dropout 
equal to $0.10$ for fully-connected layers was used during training only. 

We performed modifications at the decoder's 3rd hidden layer, since some transformation from the $z$-space should be necessary to capture underlying data invariance / structure that is then decoded to the final image. 
However, modifying lower layers also produced a variety of creative results - analysis of decoder layer choice impact is future work. 
Unless otherwise stated, results in the main paper were obtained by perturbing five neurons during decoding.  For the low-active method, we used neurons whose activations (see Method Section) were within the 1st and 15th percentiles of the neuron percent activations ($a_{j}^{k}$) for the layer.  


\textbf{Sample evaluation.}
All novelty metrics reported are averaged over 10K generated samples per method. The trained in-domain classifiers yielded test accuracy of 99.28\% for MNIST and 92.99\% for FMNIST. For human evaluation of creativity, we had 9 annotators annotate a pool of $\approx$ 500 samples per dataset (we used  agreement amongst 3 or more annotators as consensus), generated from either using  neuro-inspired creative decoding  (low-active),  baseline decoding (noisy decoding  and random activation) or regular decoding.  


\section{Results}
\begin{figure*}[]    
\centering
\includegraphics[width=6.8 in]{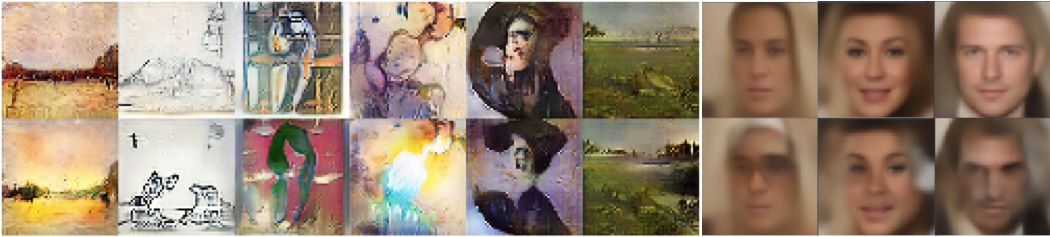}  
\caption{Results on WikiArt (with ArtGAN) and CelebA (using VAE) (top: regular, bottom: modified using low-active decoding)}
\label{fig:color}
\end{figure*}
\textbf{Human annotation.} Figures \ref{Creativity}E and \ref{Human}A present the analysis of human annotations of the generated samples by 9 annotators. For comparison, samples that were annotated as ``novel but not creative'' and ``not novel or creative'' are also shown (Fig. \ref{Human}B-C). Visually, creative samples generated from the latent space of MNIST digits do not appear digit-like anymore, but look more like symbols. In contrast, those generated from FMNIST still resemble fashion objects; however, novel objects such as a shirt with one sleeve, sock, bag with asymmetric handle, or front-slit kurta were found. Comparison with ``novel but not creative'' images confirms the known association of both novelty and value with the human perception of creativity. Interestingly, the combined dataset VAE outputs creative images that are distinct from the MNIST- and FMNIST-only cases.
The present perception of “creativity” is mostly aesthetical, so experiments on more complex ``artsy" datasets are needed; we show some results in Figure \ref{fig:color}. 

\begin{table*}[hbt!]
\small
\setlength\tabcolsep{4.5pt}
\centering
\caption{Human annotation results: L1 (Creative), L2 (Novel but not creative), L3 (Not novel or creative), L4 (Inconclusive). Values are normalized fraction of annotated instances (with a consensus of 3 or more users) within each decoding scheme. We also report average reconstruction distance, $D_{r}$,  of all generated samples by each method.
Highest L1 (creative) fraction and $D{r}$ are marked in bold.}

\begin{tabular}{l|ccccc|ccccc|ccccc}
\hline 
  & \multicolumn{5}{c|}{MNIST } & \multicolumn{5}{c|}{FMNIST } & \multicolumn{5}{c}{MNIST+FMNIST }\tabularnewline
\hline 
  & L1  & L2  & L3 & L4 & ($D_r$) & L1  & L2  & L3  & L4 & ($D_r$)   & L1  & L2  & L3   & L4 & ($D_r$)  \tabularnewline
\hline
Low-active & \textbf{0.39} & 0.50 & 0.11 & 0.00 & \textbf{4.87} & \textbf{0.27} & 0.52 & 0.14 & 0.07 & \textbf{5.49} & \textbf{0.26} & 0.54 & 0.17 & 0.04 & \textbf{4.91} \tabularnewline
Noisy & 0.17 & 0.07 & 0.76 & 0.00 & 1.78 & 0.19 & 0.09 & 0.71 & 0.01 & 1.91 & 0.25 & 0.22 & 0.51 & 0.01 & 1.97 \tabularnewline
Random & 0.24 & 0.46 & 0.30 & 0.00 & 4.21 & 0.17 & 0.62 & 0.16 & 0.06 & 4.93 & 0.19 & 0.61 & 0.15 & 0.05 & 4.68 \tabularnewline
Regular & 0.10 & 0.03 & 0.85 & 0.01 & 1.45 & 0.13 & 0.08 & 0.76 & 0.03 & 1.22 & 0.23 & 0.17 & 0.55 & 0.05 & 1.59 \tabularnewline

\hline 
\end{tabular}

\label{ranking}
\end{table*}

\textbf{Comparison with baseline methods.} 
Normalized fraction of creative samples with low subject variability (Table \ref{ranking}, L1 columns) suggests that the low-active method  constantly outperforms the baseline methods (random flipping of ``off'' neurons, noisy decoding) and regular decoding in single dataset scenario; the relative gain clearly depends on the dataset. 
Noisy decoding performs similar to regular decoding (\textit{i.e.} generates samples similar to training data), while random activation of ``off'' neurons has a higher tendency to produce novel (but not creative) samples. Therefore, the special effect of ``creative'' decoding cannot be replicated by simply flipping random ``off'' neurons (Table \ref{ranking})
- the neuro-inspired selection and flipping of low-active neurons is what promotes creativity in generations. 
Training on combined dataset enables generation of  creative samples by using baseline and regular methods as well, likely due to the extended capacity of the VAE itself (interpolating between unrelated object types). 
An interesting observation emerges  from the  average reconstruction distance ($D_r$):  The low-act method consistently yields samples with higher reconstruction distance on average for all datasets, demonstrating it's enhanced ability of generating out-of-distribution samples.   
However, those generated samples may not always be perceived as creative by human , \textit{e.g.} in the combined data scenario, as the meaningfulness might disappear.  Additionally, we ran categorical variable significance tests (chi-squared and G-tests) between our methods and baselines. Low-act method was found to be significantly different from all baselines (p<0.05) for all datasets and tests, except for the random baseline in the combined dataset scenario.

\begin{table*}[hbt!]
\small
\setlength\tabcolsep{4pt}
\centering
\caption{Comparison between human judgment  and novelty metrics: Novelty score  $NS(z)$ (considering top 5 nearest neighbors),  reconstruction distance $D_{r}$, in-domain MNIST classifier entropy $ICE_{M}$,  in-domain FMNIST classifier entropy $ICE_{F}$, in-domain score ($IS$) obtained using  one-class SVM classifier. L1 (Creative), L2 (Novel but not creative), L3 (Not novel or creative), L4 (Inconclusive).}

\small
\scalebox{1.0}[1.0] {
\begin{tabular}{l|cccc|cccc|ccccc}
\hline 
 & \multicolumn{4}{c|}{MNIST} & \multicolumn{4}{c|}{FMNIST} & \multicolumn{5}{c}{MNIST+FMNIST}\tabularnewline
\hline 
  & $NS_{z}$  & $D_{r}$  & $ICE_M$  & $IS$  & $NS_{z}$  & $D_{r}$ & $ICE_F$ & $IS$  & $NS_{z}$ & $D_{r}$  & $ICE_M$ & $ICE_F$  & $IS$\tabularnewline
\hline 


L1 & 1.305 & 5.361 & 0.215 & 419.507 & 1.378 & 5.650 & 0.126 & 341.557 & 1.429 & 4.716 & 0.251 & 0.126 & 279.086 \tabularnewline
L2 & 1.291 & 5.486 & 0.265 & 434.823 & 1.367 & 5.751 & 0.218 & 488.675 & 1.516 & 5.207 & 0.293 & 0.199 & 323.443 \tabularnewline
L3 & 1.100 & 3.084 & 0.100 & 492.340 & 1.224 & 3.667 & 0.137 & 434.923 & 1.263 & 3.284 & 0.273 & 0.169 & 415.610 \tabularnewline
L4 & 1.157 & 4.171 & 0.650 & 691.493 & 1.309 & 5.197 & 0.289 & 524.912 & 1.453 & 5.453 & 0.463 & 0.228 & 541.208 \tabularnewline 

\hline 

\end{tabular}
}

\label{humanvsml}
\end{table*}

\textbf{Relation between human judgment and novelty metrics.} Next, we report the values of novelty  metrics and their relation with the human judgment of creativity in Table \ref{humanvsml}. The std deviations (std) (not shown) are small for $Dr$ and $kNSz$ (\textit{e.g.}, std is 20\% of the average $D_r$), while ICE and IS show slightly higher variability.  Nevertheless, the overall trend of ML metrics between creative and not-creative images remains consistent across all datasets.

 Table \ref{humanvsml} reveals that a combination of  low in-domain score, and high $D_r$ and $kNS_z$ with respect to the L3 (not creative or novel)  samples are key characteristics of the creative images, on average. Same holds for novel (L2) generations; however, novel samples lie somewhere in between regular (L3) and creative (L1). These results are consistent with  earlier findings \cite{cherti2017out},  demonstrating that out-of-class metrics capture well the creative capacity of generative models. 

We further trained a ``creativity'' classifier using surrogate metrics (Table \ref{humanvsml}) as features, and found a combination of metrics provides superior predictability over any single metric.  For example, a trained L1-regularized logistic regression classifier  for predicting consensus "creative" or "not creative" on  FMNIST evaluations yields a 10-fold CV mean accuracy of 71.0\% whereas the best single-metric accuracy was 60.7\%.

In accordance with human evaluation results, the number of flipped neurons required to induce changes in surrogate metrics that are indicative of creativity was minimal for neuro-inspired low-active method  compared to  other methods (result not shown). Activating multiple ``off" neuronal clusters together was found effective as well, emphasizing need for atypical neuronal coordination for novelty generation. Linear interpolation in the $z$ space produced opposite trend in terms of surrogate metrics (in the context of creativity) and visual inspection confirmed resulting samples were not novel.
We also performed additional experiments by turning off active neurons (result not shown). To yield any significant effect required turning a larger number of neurons off. Typically, results quickly become less coherent or meaningful (e.g. become blurry), as we turn more active neurons off. This is likely due to a combination of redundancy and task- focused nature of highly active neurons. 


\textbf{Applications to more complex datasets}
Figure \ref{fig:color} shows random generations of  low-active decoding applied to ArtGAN trained on WikiArt dataset and a larger VAE trained on CelebA, which 
demonstrate generality and effectiveness of the proposed creative decoding approach.  The method works as well, if not better, on bigger networks for larger, more complex, colored images and also with GANs.  Full analysis and human evaluation of these results are left to future work.  Figure \ref{fig:color}   indicates that the proposed creative decoding causes changes in color, style, texture, and even shape - sometimes seeming to add components that fit into the scene.  \textit{E.g.}, in Figure \ref{fig:color} the first WikiArt example image seems to have an added sunrise effect, in the fourth  the middle portion is converted into a waterfall, and in the fifth  a circular shape is changed to an arm and Elvis-style hair.  For CelebA, we see sunglasses being added, shape structure / ethnicity changed, etc.

\textbf{Comparison with recent creative generation methods}
To our knowledge, the proposed method  is the first approach for generating creative modifications  that is fully unsupervised, model-agnostic (can be applied to any trained generator network) and does not require any special model training.  Earlier reports  \cite{cherti2016digits,kegl2018spurious} used either specific training or supervised learning to generate novel images. 
Since  the code or model from those studies is not publicly available to our knowledge, it is not possible to compare our results with those directly, although visual inspection reveals high similarity with some of the ``symbols" generated from MNIST in \cite{cherti2016digits,kegl2018spurious}.  We also estimated Inception Score \cite{salimans2016improved} using the MNIST classifier and found that the  creatively decoded samples have lower score (within 7-8 range) than  test samples (9.9) and significantly higher score than random samples (2.9), consistent with  \cite{kegl2018spurious}.

\section{Discussion and Future work}
Prior work has shown that  high decoder capacity enables easier posterior inference; at the same time the model becomes prone to over-fitting \cite{kingma2016improved}. 
In fact, high capacity decoders in a VAE-setting are known to ignore latent information while generating, which has been widely addressed 
\cite{kingma2016improved,rezende2015variational}. 

The presence of inactive latent units in a trained VAE decoder originates from regularization. Those low-active neurons are sparsely activated, not important for reconstruction / classification, and often encode unique sample-specific features - so are likely not part of the winning ticket \cite{frankle2018lottery}; effects of pruning them will be investigated in future.
 While our intent is not to downplay the complexity of the human brain and the creative cognition process, the fact that exploiting the extra unused capacity of a trained decoder in a brain-inspired manner provides access to novel and creative images  is interesting. Future work will include testing creativity of trainable decoders with purposely added extra capacity and exposing the model to a more complicated multi-task setting. Additionally, memory retrieval in a deep neural net by  disentangling ``low-active'' neurons  and then activating them at test time will be investigated. 
Surrogate metric design for better capturing human perception of creativity will also be investigated in future toward empowering machine learning models with  ``creative autonomy'' \cite{jennings2010developing}.



\medskip



\bibliographystyle{unsrt}

\clearpage
\section{Supplementary Material}
\subsection{Primary Datasets}
As mentioned in the main paper (see references there-in) our two primary datasets used were MNIST, a database of hand-written digits, and FMNIST, a database of fashion categories,  both contain  60000 training and 10000 test images, each image is $28 \times 28$ pixel grayscale and associated with a label from one of the 10 classes. From the training set we used 55000 images for training and 5000 for validation when training each classifier and VAE.

We also created a combined dataset - the concatenation of the MNIST and FMNIST images - to test the approaches with a generative model trained on a more diverse input space.  In this case there were a total of 110K images for training in the combined set (concatenating the 55K training sets from each dataset), and 10K for validation.  For the combined dataset, we trained a VAE on this combined train data and hold-out validation data in the same way. However, for classification, we still used the FMNIST and MNIST specific classifiers to evaluate class label outputs for the combined data (as each image was already an image from one of those two datasets).  Therefore we use and report entropy and FID with respect to each classifier (FMNIST and MNIST).  

\subsection{Decoding method details}

Here we describe our decoding modification methods with additional detail.  Recall that all of our proposed methods turn on ``off'' neurons, and that dead neurons (ones that are never active across training data) are excluded from consideration and manipulation.  Refer to Section \ref{sec:method} for definitions and preliminaries. 

\textbf{Correlation method}
The correlation method first (1) randomly selects an ``off'' neuron from the $\mathcal{D}$ group that is least-correlated with the most-activated group of neurons;  then (2) selects additional  neurons that are correlated with the selected deactivated neuron; and (3) finally turns all of them on. The idea is that, these deactivated neurons are not correlated with the most active ones, and so can be viewed as instance-specific task-negative neurons.  By using correlation, concepts encoded in multiple neurons can possibly be better captured  than pure random selection. Optionally, the same number of neurons can be turned off from those most correlated with the selected high-active neurons.  Details are provided in Algorithm \ref{alg:corr}.
\begin{algorithm}
\caption{Correlation method}
  \label{alg:corr}
\textbf{Input:} Layer output $d_{jz}^{k}$; selection fraction $\kappa$; fraction of neurons to flip on $\rho$
\begin{algorithmic}
\State  $\gamma_j \gets (1/|\mathcal{A}|)\sum_{h \in \mathcal{A}} R_{jh}^{k} d_{hz}^{k}, j \in \mathcal{D}$
\State $s \gets$ random select from smallest $\kappa |\mathcal{D}|$ neurons of $\{\gamma_j\}$
\State $\mathcal{D}_s \gets {\rm argmax}_{\mathcal{D}'\subset \mathcal{D}, |\mathcal{D}'|=\left \lfloor{\rho|\mathcal{A}|}\right \rfloor }\sum_{h \in \mathcal{D}'}R_{sh}^{k}$
\State $d_{jz}^{k} \gets o_{j}^{k}, \forall j \in \mathcal{D}_s$
\end{algorithmic}
\end{algorithm}

\textbf{Clustering method} 
The clustering method follows a similar approach as the correlation method, except instead of considering individual neurons, clusters of neurons are considered, which represent instance-specific task-negative sub-networks.  Spectral clustering 
is applied to the layer output correlation matrices from the training data ($R^k$) to get the cluster memberships.  For a given instance and decoder layer, one-or-more clusters with lowest percent activation are randomly selected, where percent activation for cluster $\mathcal{C}^k$ is $(1/|\mathcal{C}^{k}|)\sum_{j \in \mathcal{C}^k} \mathbbm{1}_{d_{jz}^{k} > \tau}$. The percent activations of the selected clusters  are then increased by randomly turning on more neurons in those clusters until the specified number of neurons to turn on (or alternatively percent increase) is reached.

\textbf{``Low-active'' method} 
The ``low-active'' method (Algorithm \ref{alg:low-act}) captures the task-negative concept in a more realistic sense - by identifying those neurons that typically have low activation across all the training data and turning some fraction of them on at test time.  Specifically, a neuron is selected from  deactivated neurons that have the lowest percent activations ($a_{j}^{k}$). Next, ``low-active'' neurons that are  strongly  correlated  with the selected neuron are turned on.

\textbf{Non-specific ``low-active''  method} 
This method extends the low-active method to consider the activation of neurons across different classes - to select activating neurons that are originally overall low-active (across the entire training data) and do not show activation  beyond a threshold  for any specific class - \textit{i.e.}, non-specific, as this scheme mimics more closely  the concept of the task negative default network in the brain.  
For MNIST and FMNIST data we used the class labels, but for the FMNIST+MNIST combination data, we used a label indicating which dataset the image is from (FMNIST or MNIST).

Specifically, this method uses two criteria for selecting neurons.  (1) that the max percent activation across all training data in any given class is below a threshold (as in low-active method but for each class) and (2) that the entropy of percent activation across classes is below a threshold.  In this study, we used  maximum percent activation of 15\% and required the activation of the neuron to be of high entropy (in the top 30\% of entropy values for the neurons in the layer).  For the one-class SVM classifier, we used an RBF kernel with gamma $=0.1$ and 15\% outlier fraction.

\textbf{Complexity} 
Another advantage of our proposed approaches is they introduce little-to-no overhead during the decoding process, as properties of neurons (such as correlations, cluster membership, low-active status, etc.) are all pre-computed.  Therefore during the normal decoding process it only becomes necessary to operate over the set of non-dead neurons in the specified layer and their outputs $d_{jz}^{k}$.  Therefore complexity is $O(p)$ for each operation across neurons (such as random selection or filtering to group membership or criteria), where $p$ is the number of alive neurons in a layer being modified during decoding.  However, some methods choose most or least correlated neurons, so at worse it could amount to sorting the alive neurons in the layer, $O(p\log{}p)$.  Therefore modified decoding complexity for all methods is at worst $O(p\log{}p)$, assuming a small fixed number of neurons selected to be turned on.

\subsection{Classifier and VAE details}

For F/MNIST VAEs the encoder network consisted of 3 fully- connected layers (1000, 500, 250), and for CelebA (dataset details in Section \ref{sec:celeba}), 9 convolutional layers (3 groups with 32, 64, then 128 filters - each having 2 3 x 3 kernels then 1 2 x 2, with strides 2, 2, and 1) followed by 2 fully-connected layers (1000, 500). After training, number of ``dead neurons'' for layer 1, 2, and 3 were (0,132,127), (0, 136, 226), (0, 52, 87) for MNIST, FMNIST, and MNIST + FMNIST , respectively. Number of ``off'' neurons varies for each sample. For layer 2, the mean and std. dev. were 792 $\pm$ 15,708 $\pm$ 14, and 797 $\pm$ 32 for MNIST, FMNIST, and MNIST + FMNIST.

The classifiers consisted of 5 convolutional layers (3 layers of 32 filters with 2 3x3, and 1 2x2 kernels having strides 1, 1, and 2; then 2 with 64 filters, 1 3x3 then 1 2x2 with strides 1 and 2) followed by one fully connected (200) 
layer and one output layer, with RELU activations and a dropout of $0.10$.     Classifier test accuracy was 99.28\% for MNIST and 92.99\% for FMNIST.

For both the classifiers and VAE training, we used the ADAM optimizer 
using a batch size of 128, 200-400 epochs with best validation score model kept, re-starts from best used after 10 epochs of no improvement, decreasing step size when validation error failed to decrease for 20 epochs, and the best validation set scoring model tracked and used at the end as the final model.  For the combined dataset we scaled the reconstruction negative log-likelihood loss by 1000 and for CelebA by 1e7, in order to see clearer images generated (without weighting this term more heavily generated outputs were extremely blurry in these cases / indistinguishable).

For the VAE, the latent dimension size was set to 50 for F/MNIST, 128 for CelebA, and 100 for the FMNIST + MNIST combined dataset.
The final test loss, KL-divergence, and negative log-likelihood reconstruction loss are given in Table \ref{tab:vae_losses}.

\begin{table}[h]
    \centering
    \caption{VAE test losses}

    \begin{tabular}{c|c|c|c}
         dataset & overall loss & NLL & KL-loss \\\hline
         mnist & 110.0 & 96.3 & 13.7 \\
         fmnist & 241.6 & 230.8 & 10.8 \\
         combined & 146002.9 & 145969.7 & 33.2 \\
         celebA & 75837.2 & 75259.5 & 577.7
    \end{tabular}
    \label{tab:vae_losses}
\end{table}

\subsection{Additional novelty metrics: Fr\'{e}chet Inception Distance (FID).}
FID \cite{heusel2017gans} provides an alternative approach that requires no labeled data. The samples are first embedded in some feature space (a specific layer) of a trained classifier and  a continuous multivariate Gaussian is fit to the embedding layer. Then, mean and covariance is estimated for both the generated and training data. The Fr\'{e}chet distance between these two Gaussians is then computed as the Fr\'{e}chet Inception Distance or $FID (x, g) = ||\mu_{x} -  \mu_{g} ||_{2}^2 + Tr(\sum_{x} + \sum_{g} - 2(\sum_{x} \sum_{g} )^{\frac{1}{2}} )$, where ($\mu_{x}$,$\sum_{x}$),
and 
($\mu_{g}$,$\sum_{g}$) are the mean and covariance of the sample embeddings from the data distribution and model distribution, respectively.  Lower FID means smaller distances between synthetic and real distributions.
FID is found to be sensitive to  addition of spurious modes, to mode dropping, and is consistent with human judgment, while may not capture ``memory overfitting''  \cite{lucic2018gans}. 

\subsection{``Creativity Filter''}
As explained in the main paper, as part of the human evaluation, for a given modified generation method we also randomly mixed in a portion of generated examples that passed a ``creativity'' filter (in addition to generated examples from that method without any filter).  
The ``creativity filter'' was set to require that input and latent space nearest-neighbor distances and reconstruction distance had to be greater than an interquartile range outlier threshold for regular generation distances (from a 20K sample of regular VAE generation) and also have classifier entropy greater than the mean entropy for test data but less than the mean entropy for random images (as too much entropy could indicate garbage as opposed to creativity).  The filter distance thresholds are given by $Q_2 + 1.75(Q_3-Q_1)$ where $Q_1$, $Q_2$, and $Q_3$ are 1st, 2nd, and 3rd quartiles, respectively, of distance metric values on the training data.

\subsection{Additional experiments \label{sec:celeba}}

\begin{figure}[t]
\centering
\begin{center}
\centering
\centerline{\includegraphics[width=0.5\columnwidth]{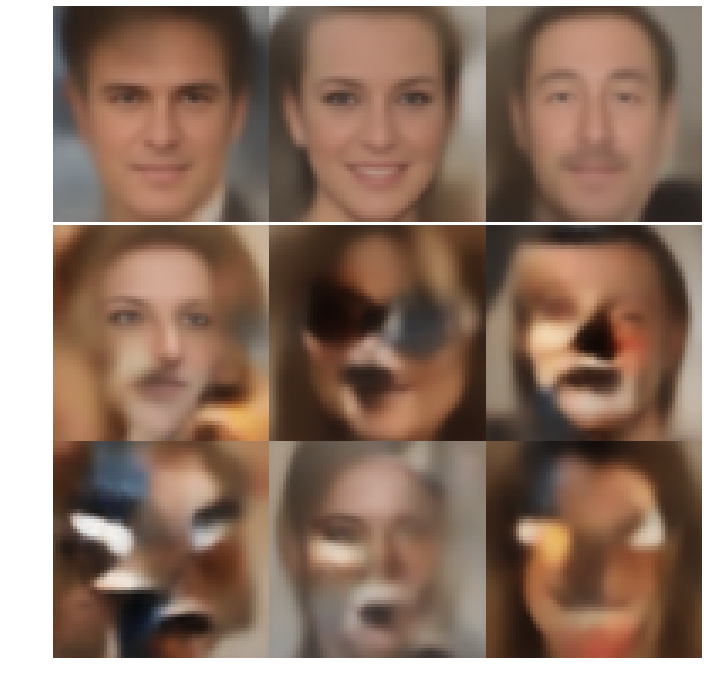}}
\centerline{\includegraphics[width=0.9\columnwidth]{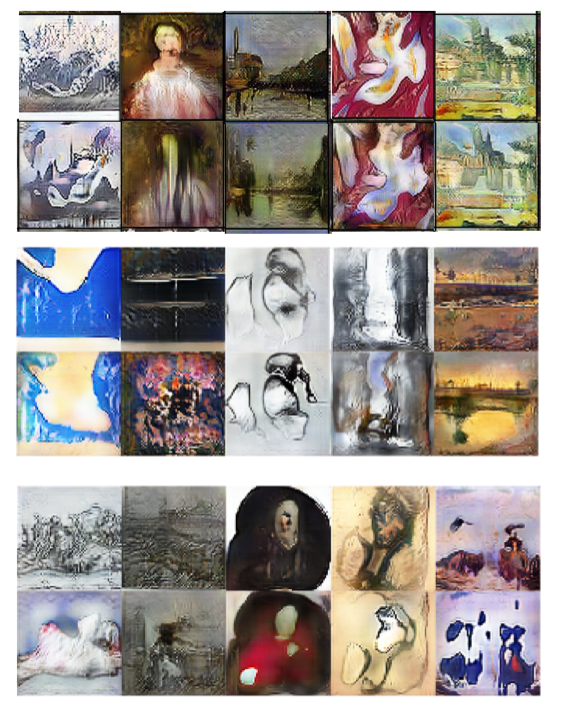}}
\caption{\textbf{Top.}CelebA images obtained by activating default TN neurons (with low-active method). Top row: normal generation from VAE.  Middle and bottom row: modified generation, using a variety of different algorithm settings for more or less extreme creative modification. \textbf{Bottom.} Examples of WikiArt images generated by low active decoding (in each panel top: regular, bottom: modified. }
\label{fig:celeba_ex}
\end{center}
\end{figure}

\textbf{CelebA experiment.} We also applied the proposed approach to celebrity face images, the CelebA \cite{liu2015faceattributes} dataset.   CelebA is a database of roughly 200,000 aligned and cropped $178 \times 218$-pixel RGB images of celebrity faces, re-sized to $64\times64$ (200599 training and 2000 test).  Although we only report preliminary results here (we do not have the exhaustive evaluation analyses of the results as for the other data sets and more analyses of the decoding settings are needed), we still want to present the early results of creative decoding on CelebA.   In this case, for ``turning on'' a neuron, setting its value to 2-times its maximum value seen in training data seems to often give too extreme results.  We find instead using a smaller scale factor such as 1 (i.e., simply setting to the max training value), or using the mean value with a scale factor greater than 1 instead works better (visual inspection).   In addition to the examples shown in the main paper, in Fig. \ref{fig:celeba_ex} we present generated CelebA samples by using the low-active decoding.  Note that these are generated with a variety of different algorithm settings, in particular varying the flip type (whether to set to a maximum activation or mean activation value) and flip scale (how much to scale this maximum or mean activation value when setting the neuron to this value).  This is to show the impact of a variety of different settings - in this way we can generate more or less extreme modified / creative results.  In this figure, the 3 images in the top row are normal generated faces, and the remaining 6 images  are various faces generated using the modified decoding method, with different settings.   

In general we see a variety of transformations and distortions with the CelebA data- with stronger perturbations (setting neurons to larger activation values when turning them on or turning on more neurons) we see more visually extreme changes, such as blurring and distorting of features and distinct colors coming out, and larger spots of color (black and otherwise) in the faces.  With subtler changes when turning on neurons, such as turning on fewer neurons,  smaller neuron fraction from a less-active cluster, or turning neurons on to less-extreme level, we see more subtle changes and distortions in the images and many correspond to specific features.  For example, we may see parts of the face rotating, or glasses being added on one or more eye, or male or female characteristics changing, or various facial hair or tinting appearing. 

Our cluster-based method seemed to provide the most control over the effects of the modification in this case, making it easier to control the effect of the modification by controlling how much of the ``task-negative'' clusters were activated, and also because different clusters tended to correspond to having different effects (such as face rotation, aging, etc.).

\textbf{WikiArt with ArtGAN experiment.}
We also experimented with applying our methods to the WikiArt dataset - a collection of images of different artwork, organized by artist, genre, and style (\url{https://www.wikiart.org}).  Specifically we used the pre-trained GAN model, ArtGAN, obtained from the authors' website (\url{https://github.com/cs-chan/ArtGAN/tree/master/ArtGAN}), trained on the Genre images.  This is an 8-layer conditional Generative Adversarial Network (GAN) generative model, conditioned on the genre; with a typical up-sampling, convolution, batch-normalization, (leaky) RELU pattern followed in the layers.  It was trained to generate 128x128 images of the art images.  Full details of the network, the data, and the training procedures can be found on the website.

This was a case of testing our method both on a richer more complex set of images as well as on GAN models for the first time (as opposed to just VAEs).  Similarly as with other datasets, we applied our method to the third layer of the generator (decoder) of the GAN - as explained previously since some amount of transformation from the $z$-space should be necessary to capture the underlying data invariance and structure that is then transformed to the final image, it makes sense not to use too-early layers, but using too-late layers would be too low-level to capture high-level, cross-image structure.

We found our creative decoding method to work well for this data and GAN architecture as well, and we show an example of creative generation results in Figure \ref{fig:color} in the main paper, for a variety of different art genres.  


\subsection{Human annotation experiment}
\begin{figure*}
\centering
\includegraphics[width=1\linewidth]{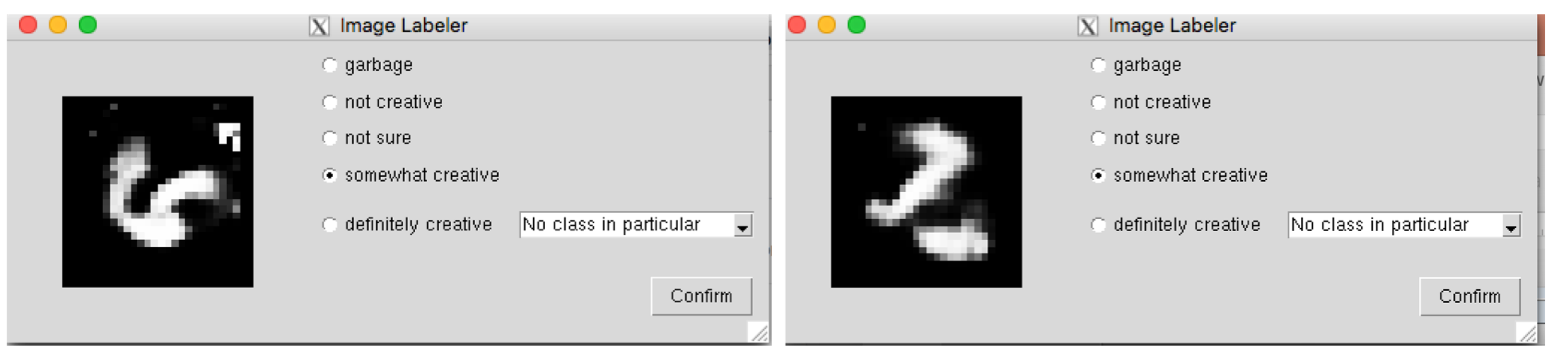}  
\caption{Snapshots of the human annotation tool}
\label{snapshot}
\end{figure*}

\begin{figure*}
\centering
\includegraphics[width=1\linewidth]{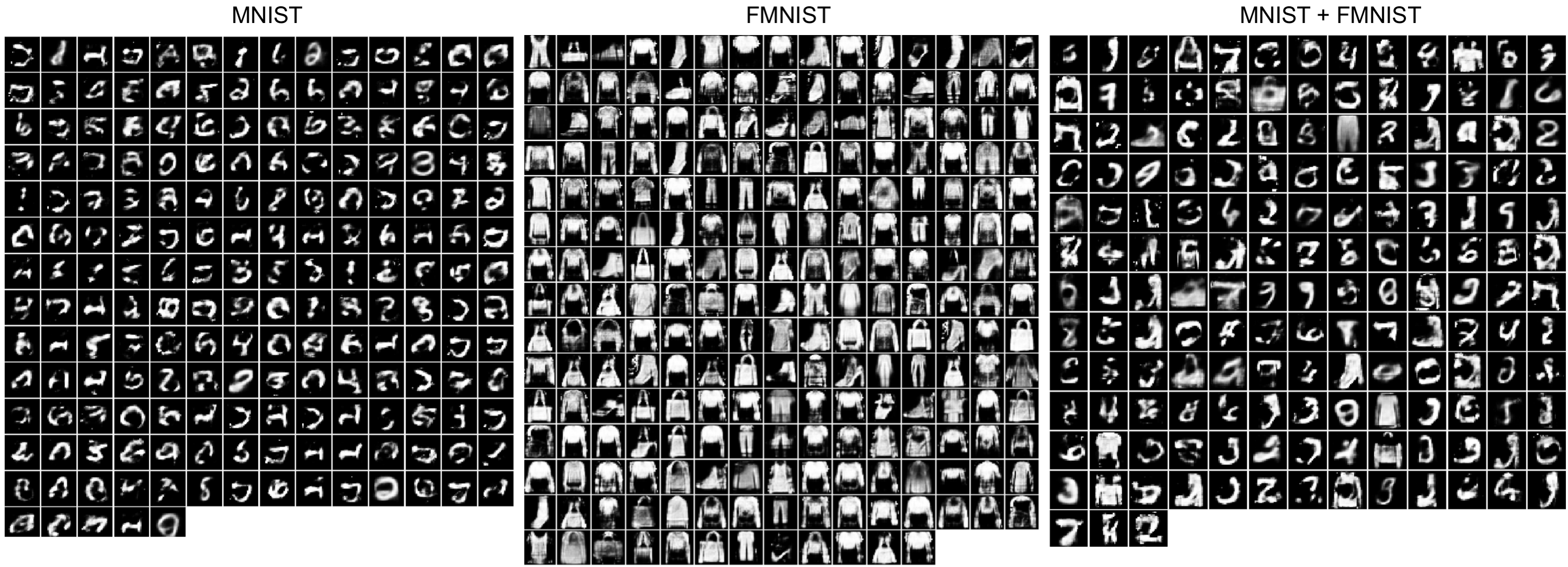}  
\caption{Human-annotated creative  images}
\label{add_creative}
\end{figure*}

\begin{figure*}
\centering
\includegraphics[width=0.5\linewidth]{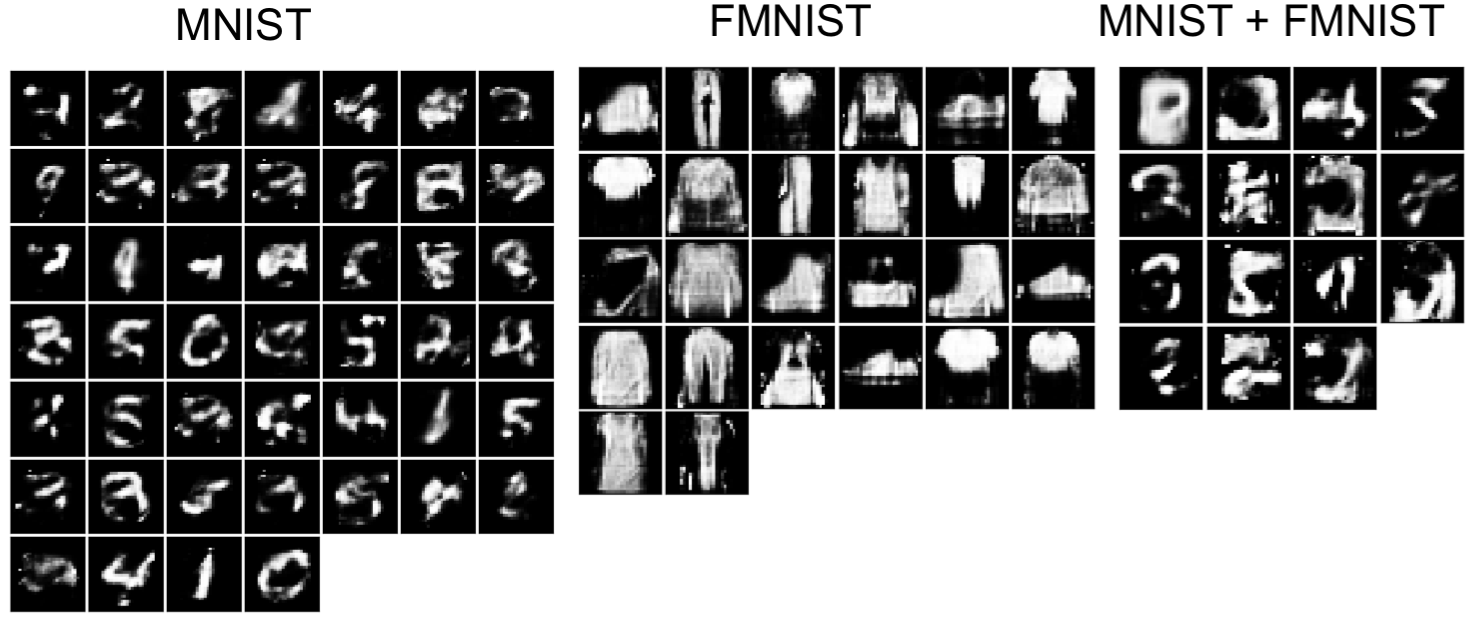}  
\caption{Human-annotated not sure images}
\label{addl_notsure}
\end{figure*}

Figure \ref{snapshot} shows snapshots of the human annotation tool. Additional examples of creative samples and not-sure samples are shown in Figures \ref{add_creative} and \ref{addl_notsure}.

\subsection{Human annotation results including the non-specific low-active method}
Here we show the complete human annotation result table, including results for the non-specific low-active method.  We excluded from the main text for simplicity, ease-of-understanding, and for space constraints.  The results are shown in Table \ref{ranking_with_non_spec}.  We see that the non-specific method mostly follows the same trend and performance as the low-active one, but the low-active overall does better in terms of creative fraction and reconstruction distance.  The non-specific version is more negatively affected by the data diversity - likely due to the entropy element causing more distant generated objects from the known ones - so that there were more rated as novel but not creative by the human users, instead of creative.

\begin{table*}[hbt!]
\small
\setlength\tabcolsep{4.5pt}
\centering
\caption{Human annotation results: L1 (Creative), L2 (Novel but not creative), L3 (Not novel or creative), L4 (Inconclusive). Values reported are normalized fraction of annotated instances (with majority vote)) within each decoding scheme. We also report average reconstruction distance, $D_{r}$,  of all generated samples by each method inside parentheses. 
Highest L1 (creative) fraction and $D{r}$ are marked in bold.}

\begin{tabular}{l|ccccc|ccccc|ccccc}
\hline 
  & \multicolumn{5}{c|}{MNIST } & \multicolumn{5}{c|}{FMNIST } & \multicolumn{5}{c}{MNIST+FMNIST }\tabularnewline
\hline 
  & L1  & L2  & L3 & L4 & ($D_r$) & L1  & L2  & L3  & L4 & ($D_r$)   & L1  & L2  & L3   & L4 & ($D_r$)  \tabularnewline
\hline
Non-specific & 0.31 & 0.52 & 0.16 & 0.02 & 4.47 & 0.23 & 0.48 & 0.15 & 0.14 & 4.50 & 0.17 & 0.68 & 0.07 & 0.08 & 4.52 \tabularnewline
Low-active & 0.39 & 0.50 & 0.11 & 0.00 & 4.87 & 0.27 & 0.52 & 0.14 & 0.07 & 5.49 & 0.26 & 0.54 & 0.17 & 0.04 & 4.91 \tabularnewline
Noisy & 0.17 & 0.07 & 0.76 & 0.00 & 1.78 & 0.19 & 0.09 & 0.71 & 0.01 & 1.91 & 0.26 & 0.22 & 0.51 & 0.01 & 1.97 \tabularnewline
Random & 0.24 & 0.46 & 0.30 & 0.00 & 4.21 & 0.17 & 0.62 & 0.16 & 0.06 & 4.93 & 0.19 & 0.61 & 0.15 & 0.05 & 4.68 \tabularnewline
Regular & 0.10 & 0.03 & 0.85 & 0.01 & 1.45 & 0.13 & 0.08 & 0.76 & 0.03 & 1.22 & 0.23 & 0.17 & 0.55 & 0.05 & 1.59 \tabularnewline

\hline 
\end{tabular}

\label{ranking_with_non_spec}
\end{table*}

\subsection{Complete analysis of novelty metrics with varying number of neurons turned on for MNIST, FMNIST, and MNIST+FMNIST and more creative decoding approaches.  \label{sec:all_metrics}}


\begin{figure}[ht]
    \centering
    \begin{subfigure}[b]{0.32\linewidth}
        \centering
        \includegraphics[width=\linewidth]{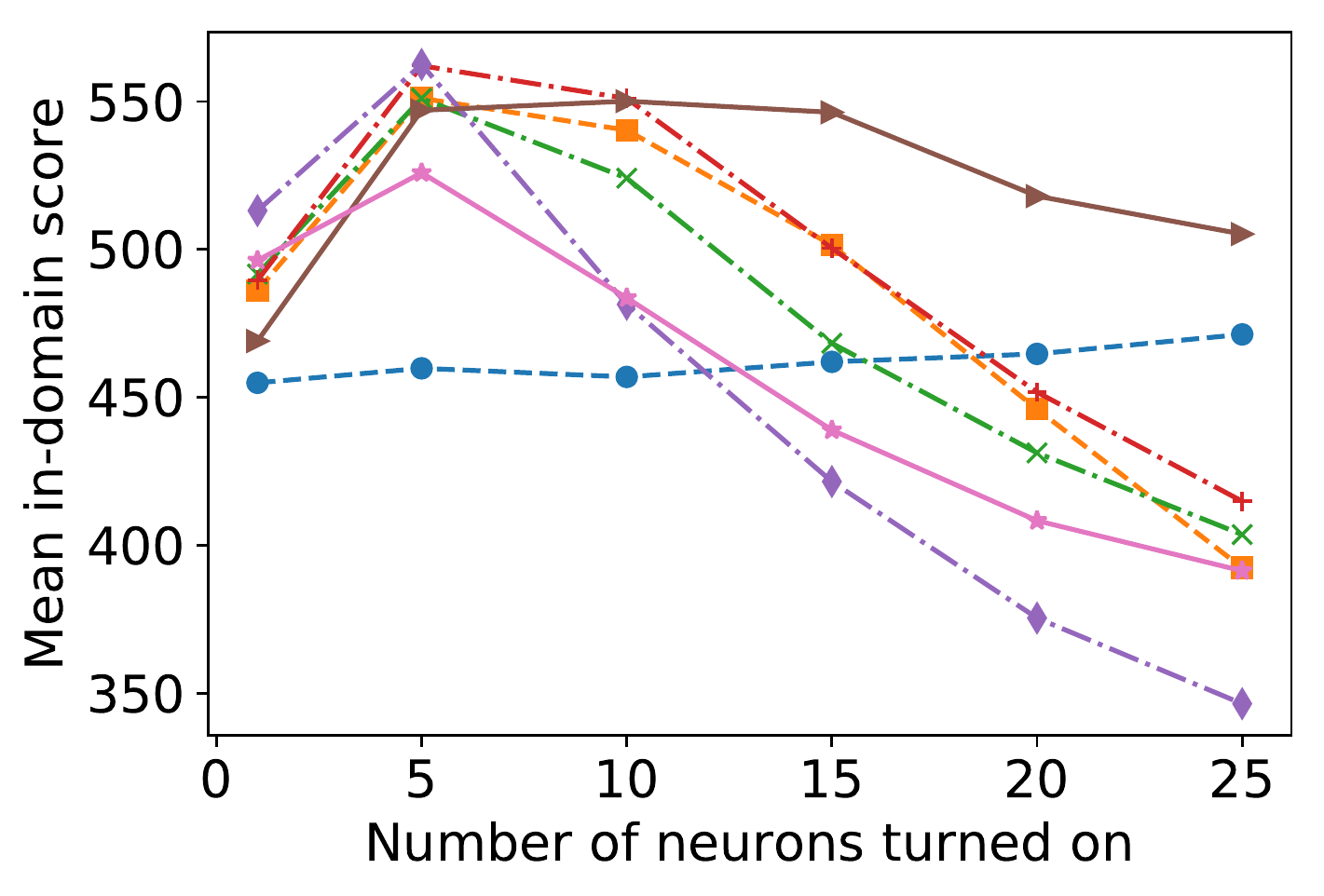}
        \caption{MNIST}
    \end{subfigure}%
    \begin{subfigure}[b]{0.32\linewidth}
        \centering
        \includegraphics[width=\linewidth]{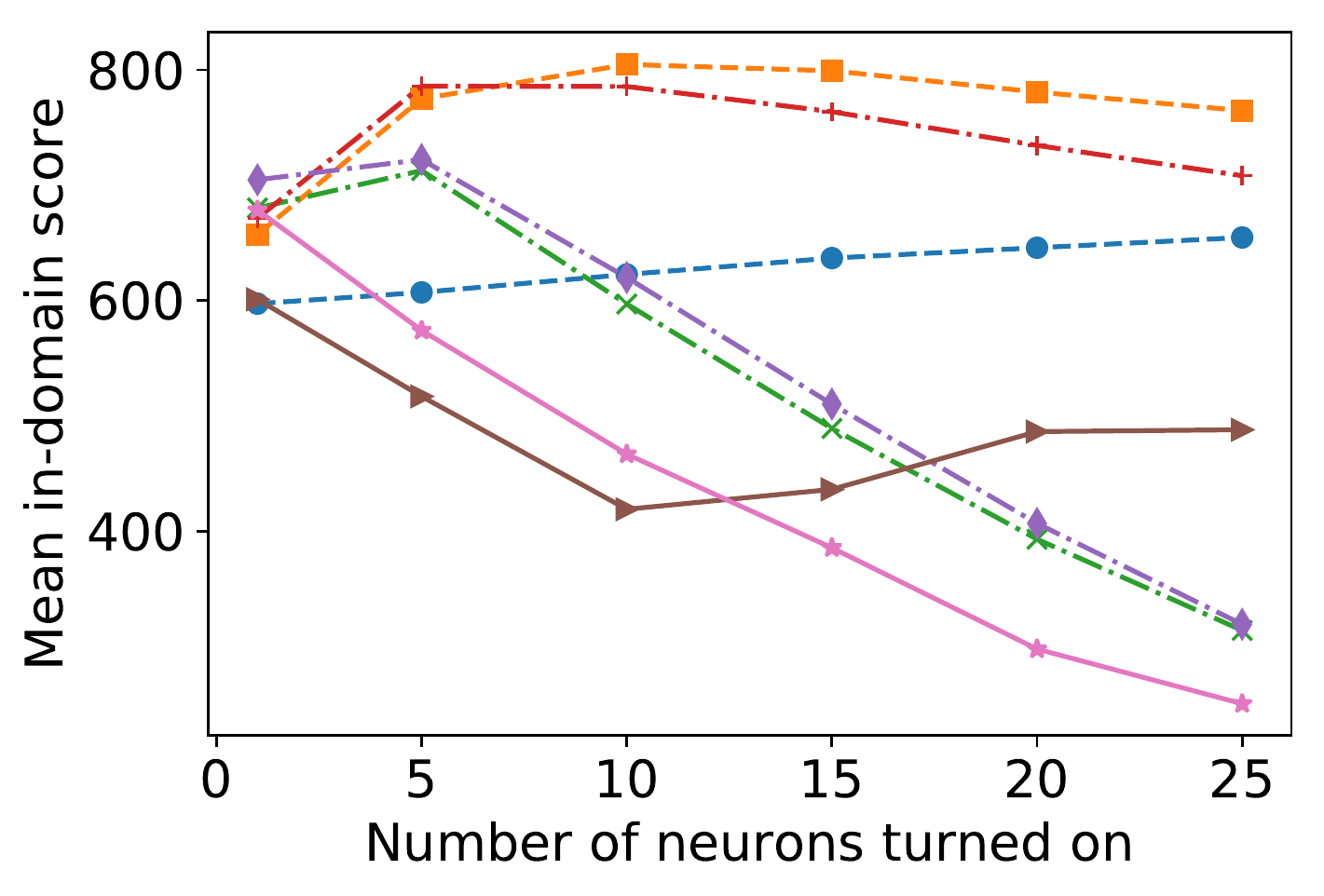}
        \caption{FMNIST}
    \end{subfigure}%
    \begin{subfigure}[b]{0.32\linewidth}
        \centering
        \includegraphics[width=\linewidth]{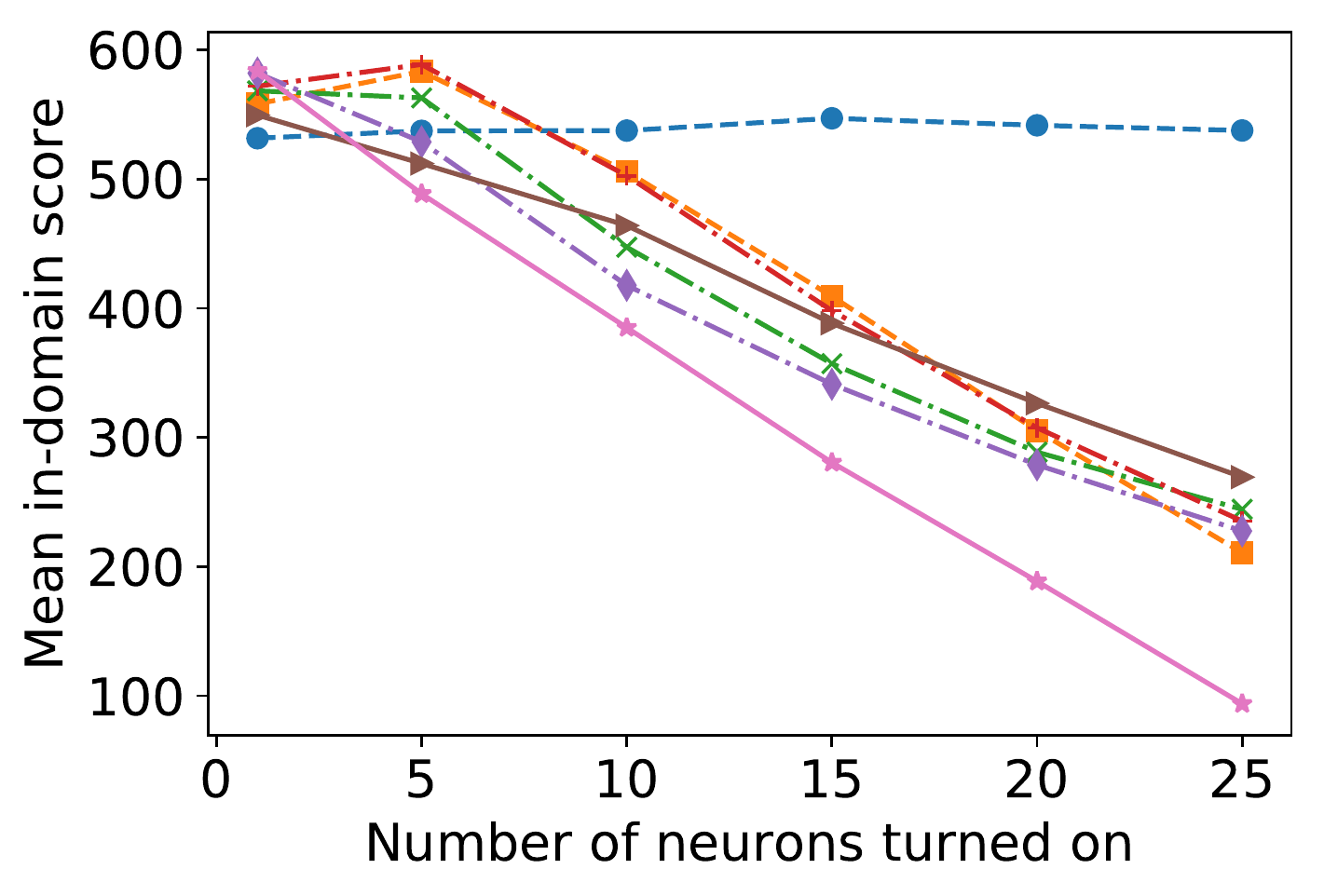}
        \caption{Combination}
    \end{subfigure}%
    \\
    \begin{subfigure}[b]{0.5\linewidth}
        \centering
        \includegraphics[width=\linewidth]{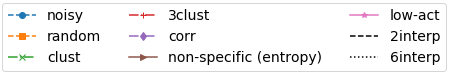}
        \caption{Legend}
    \end{subfigure}%
    \caption{Mean in-domain score from one-class classifier for each data set and approach, for increasing number of neurons turned on, from 1 to 25.  Interp. method results were excluded because their scores were significantly higher (see below tables for details).}
   \label{fig:mean_outlier_score_per_nto}
\end{figure}

\begin{figure}[ht]
    \centering
    \begin{subfigure}[b]{0.32\linewidth}
        \centering
        \includegraphics[width=\linewidth]{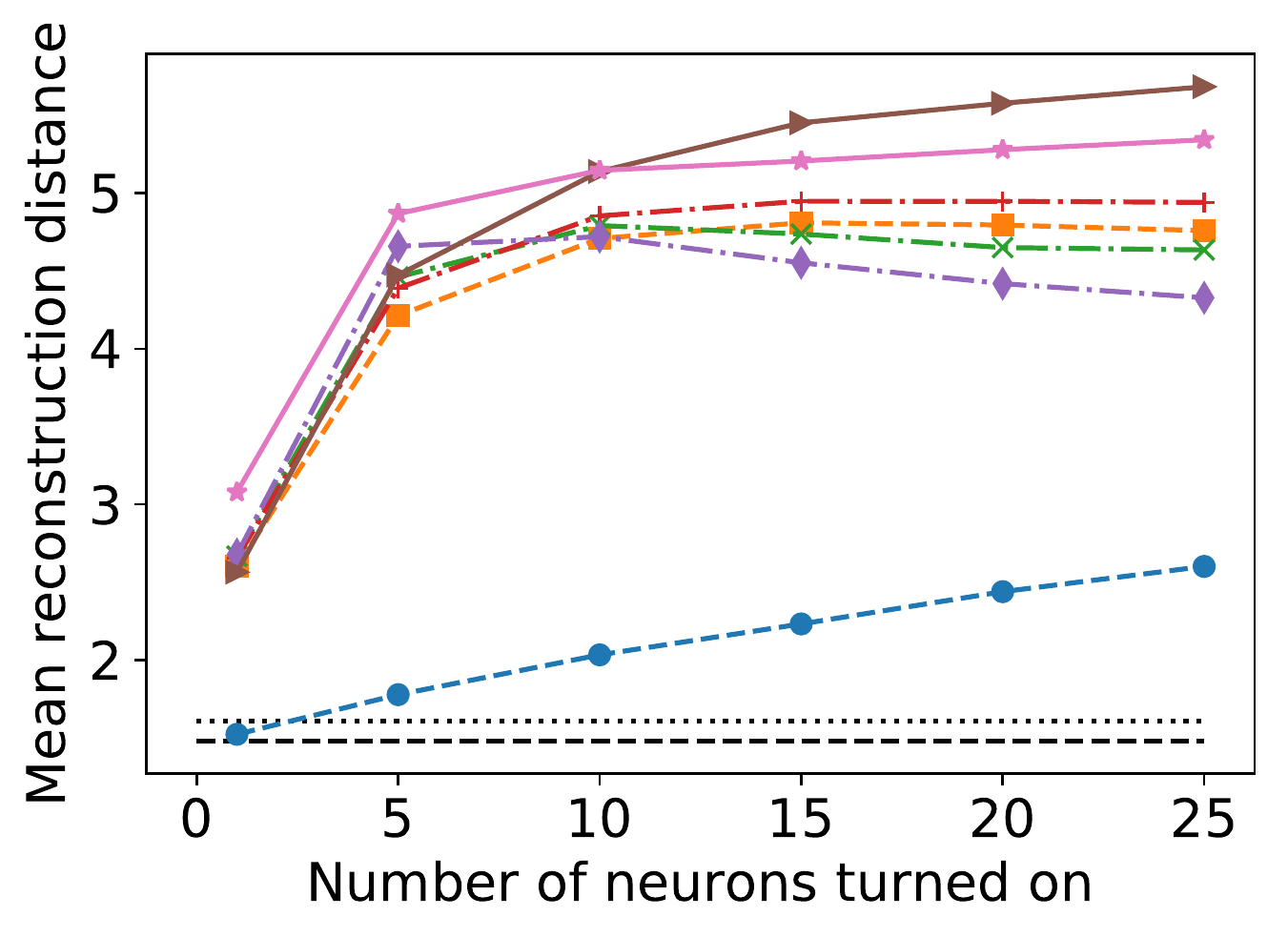}
        \caption{MNIST}
    \end{subfigure}%
    \begin{subfigure}[b]{0.32\linewidth}
        \centering
        \includegraphics[width=\linewidth]{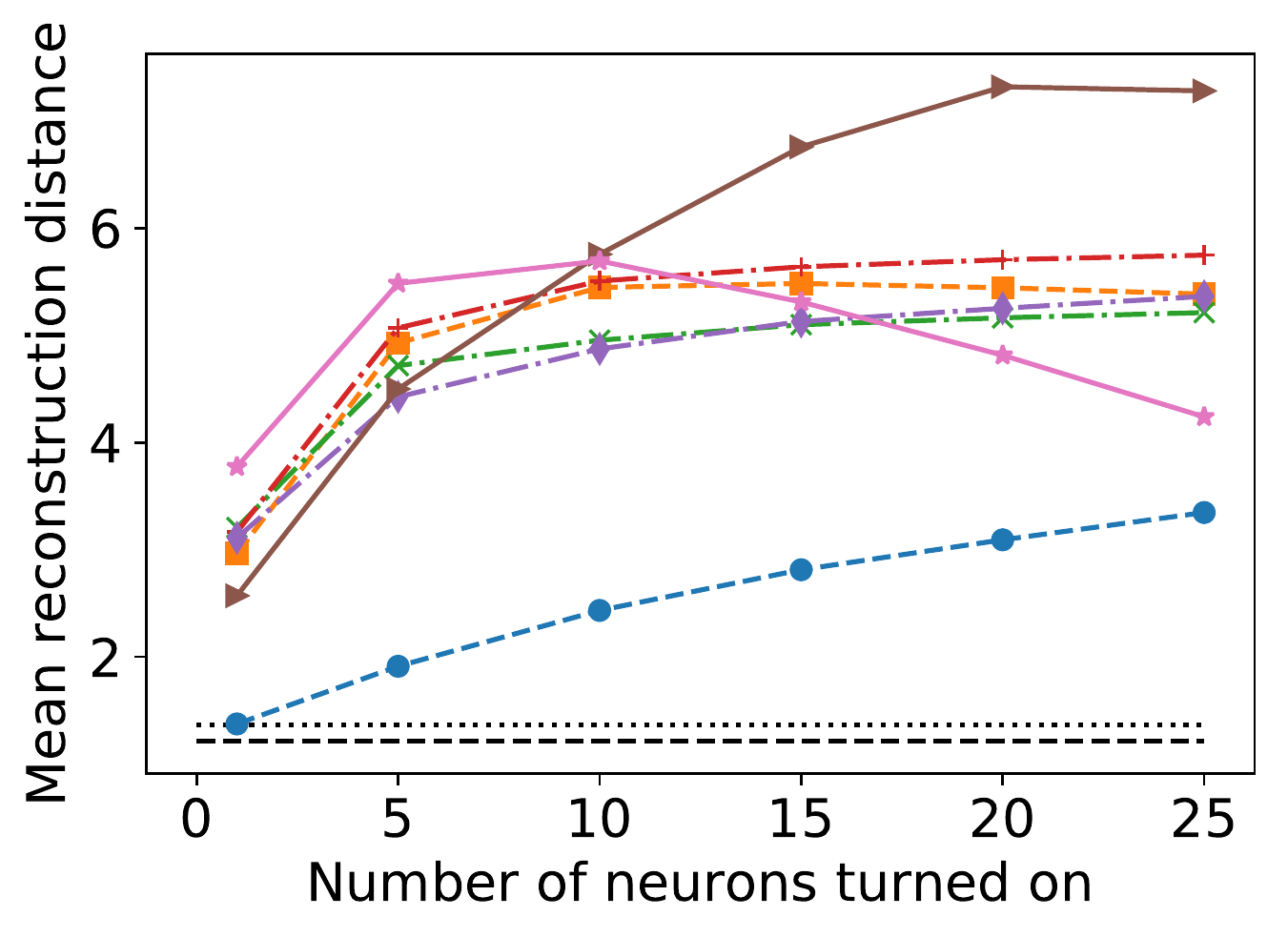}
        \caption{FMNIST}
    \end{subfigure}%
    \begin{subfigure}[b]{0.32\linewidth}
        \centering
        \includegraphics[width=\linewidth]{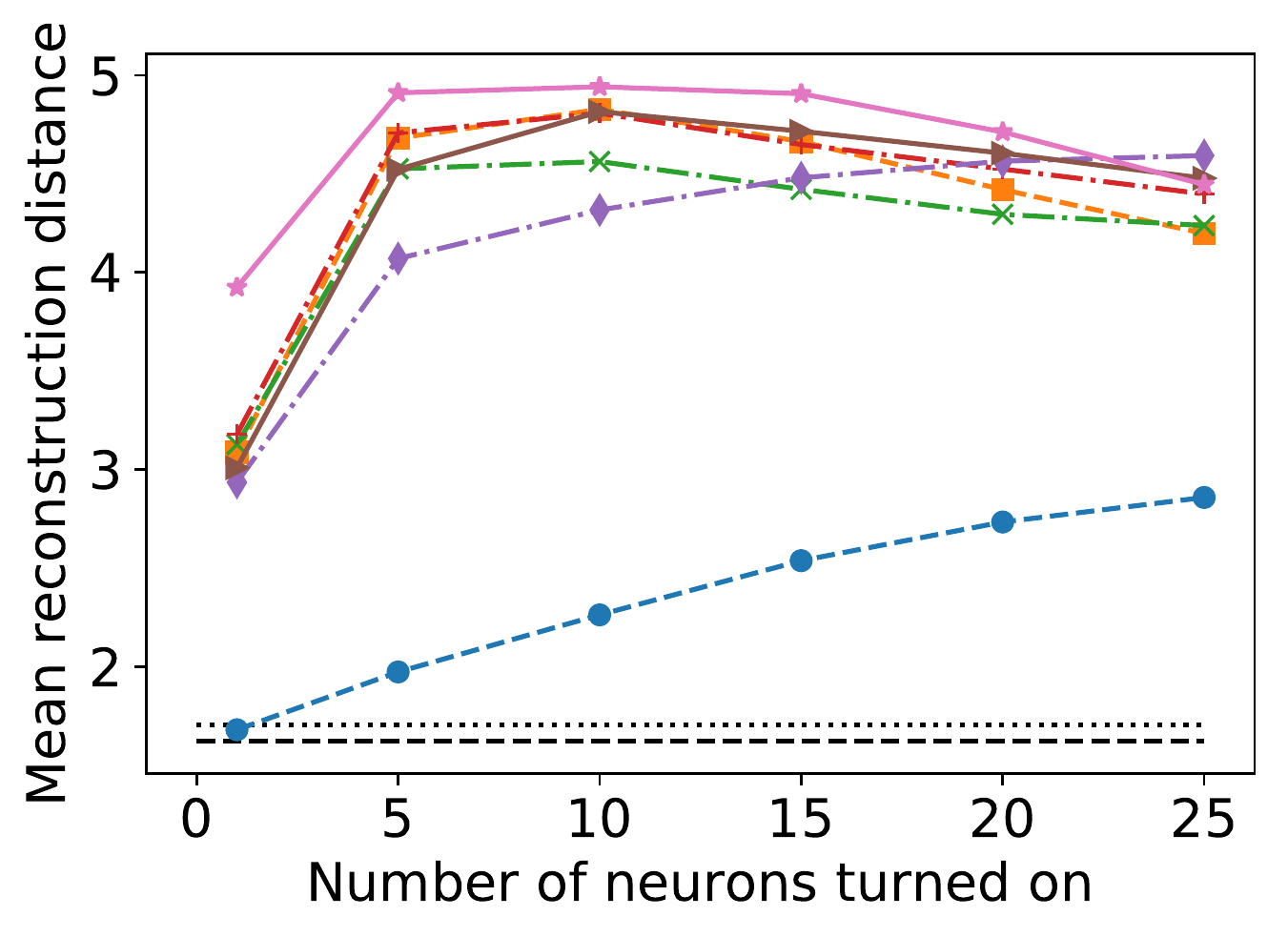}
        \caption{Combination}
    \end{subfigure}%
    \\
    \begin{subfigure}[b]{0.5\linewidth}
        \centering
        \includegraphics[width=\linewidth]{figs/legend.png}
        \caption{Legend}
    \end{subfigure}%
    \caption{Mean reconstruction for each data set and approach, for increasing number of neurons turned on, from 1 to 25.}
   \label{fig:mean_recon_dist_per_nto}
\end{figure}

\begin{figure}[ht]
    \centering
    \begin{subfigure}[b]{0.32\linewidth}
        \centering
        \includegraphics[width=\linewidth]{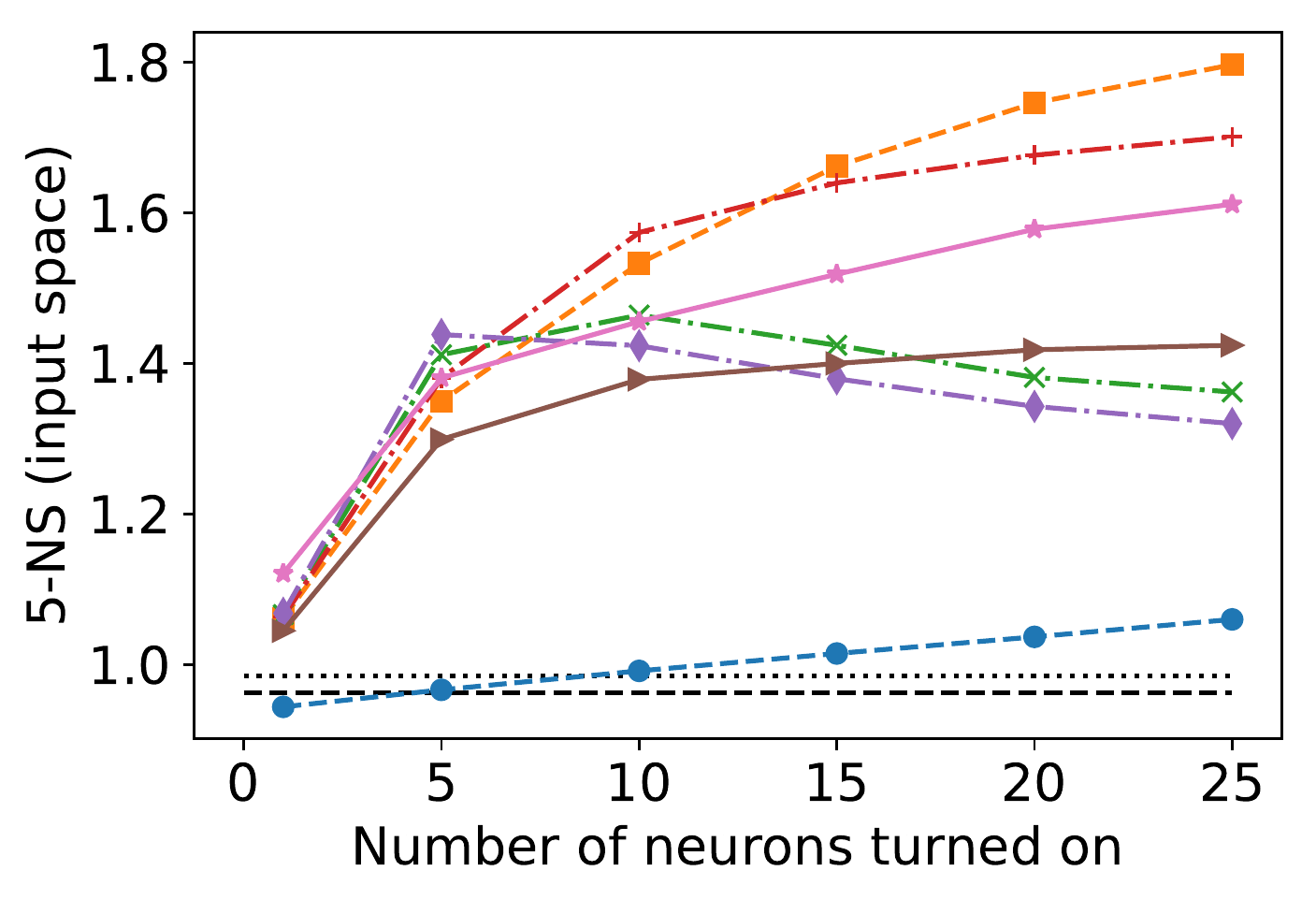}
        \caption{MNIST}
    \end{subfigure}%
    \begin{subfigure}[b]{0.32\linewidth}
        \centering
        \includegraphics[width=\linewidth]{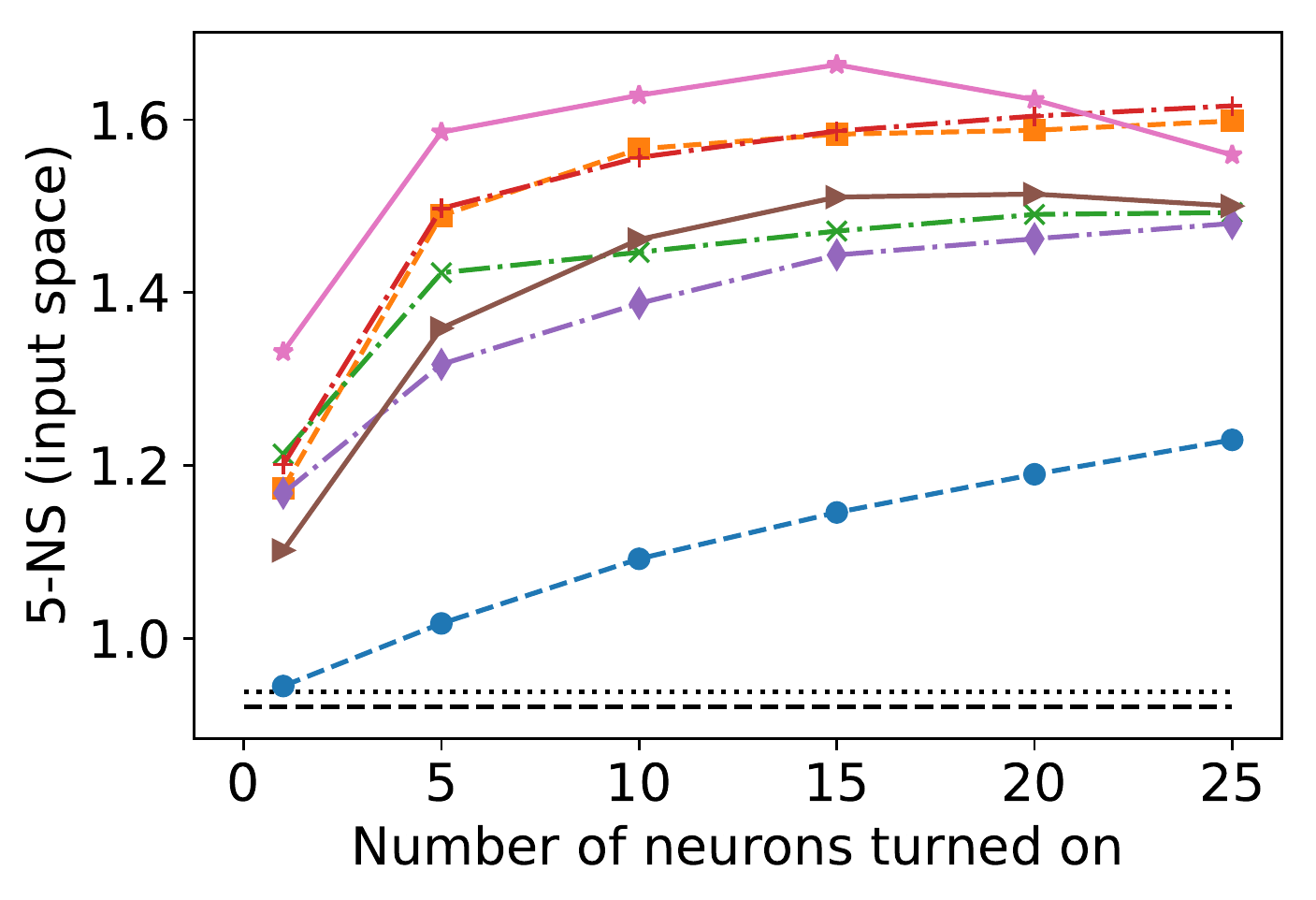}
        \caption{FMNIST}
    \end{subfigure}%
    \begin{subfigure}[b]{0.32\linewidth}
        \centering
        \includegraphics[width=\linewidth]{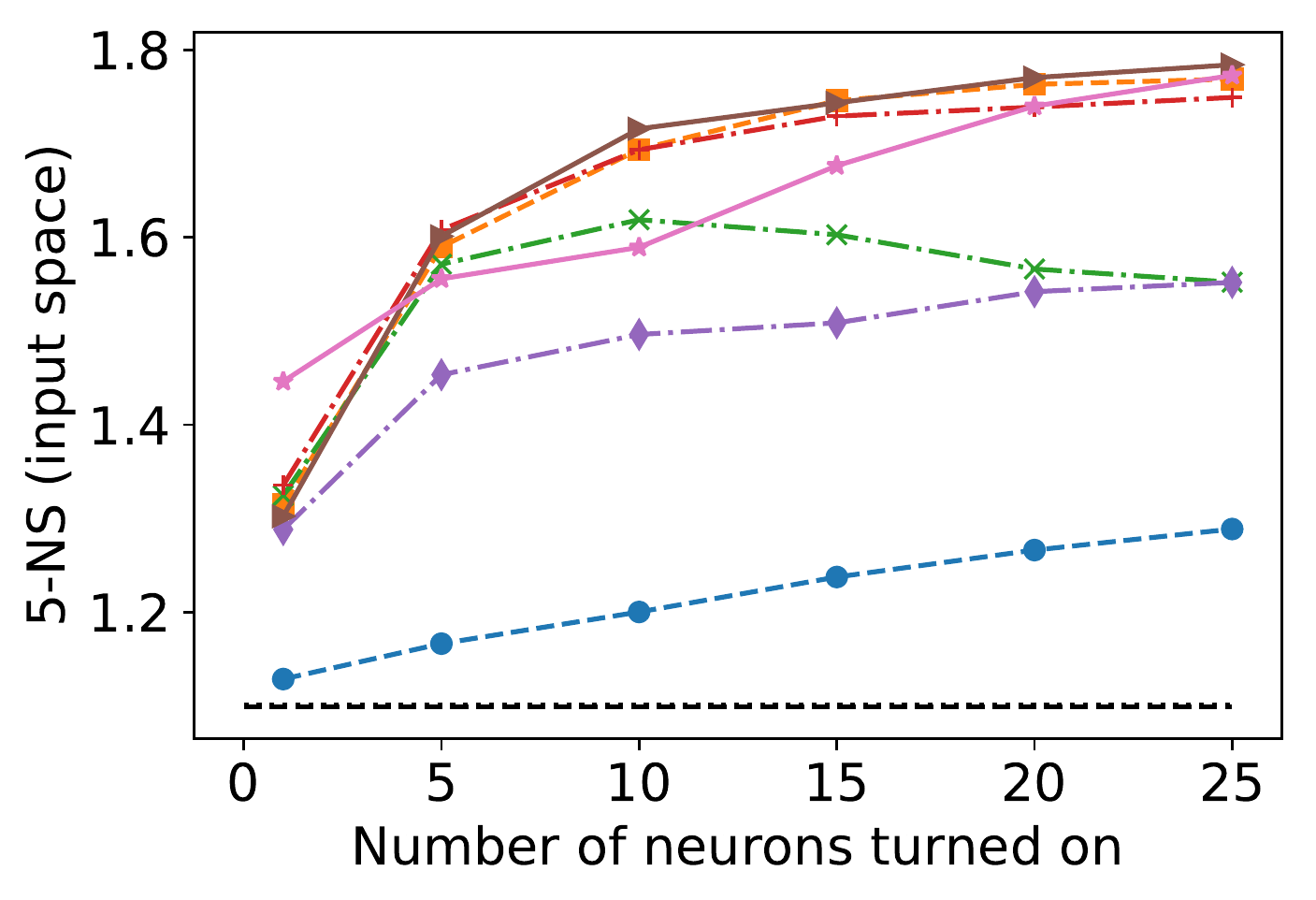}
        \caption{Combination}
    \end{subfigure}%
    \\
    \begin{subfigure}[b]{0.5\linewidth}
        \centering
        \includegraphics[width=\linewidth]{figs/legend.png}
        \caption{Legend}
    \end{subfigure}%
    \caption{Mean 5-NS (normalized $5^{th}$ nearest neighbor distance) in the input space for each data set and approach, for increasing number of neurons turned on, from 1 to 25.}
   \label{fig:mean_norm5nn_dist_per_nto}
\end{figure}

\begin{figure}[ht]
    \centering
    \begin{subfigure}[b]{0.32\linewidth}
        \centering
        \includegraphics[width=\linewidth]{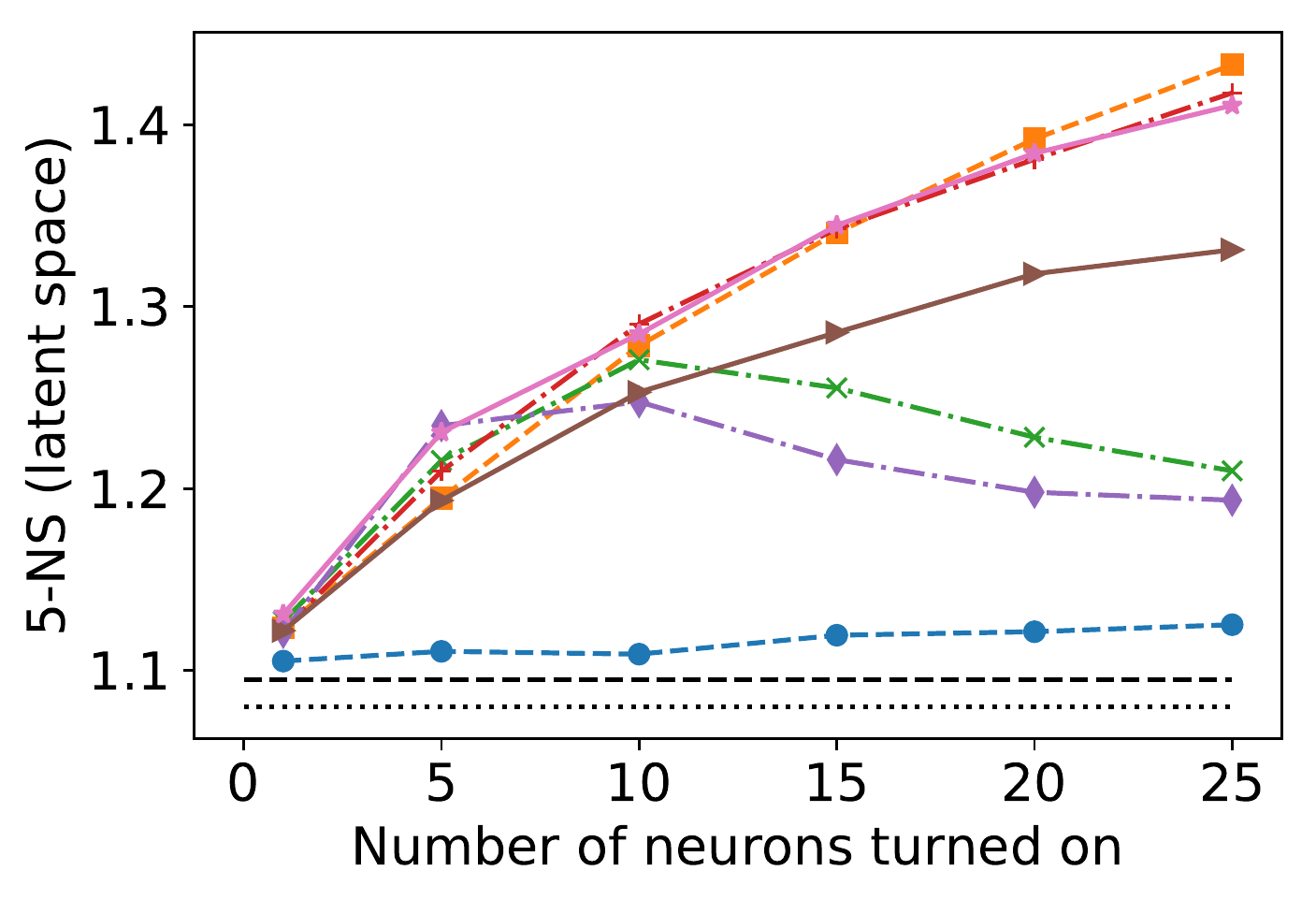}
        \caption{MNIST}
    \end{subfigure}%
    \begin{subfigure}[b]{0.32\linewidth}
        \centering
        \includegraphics[width=\linewidth]{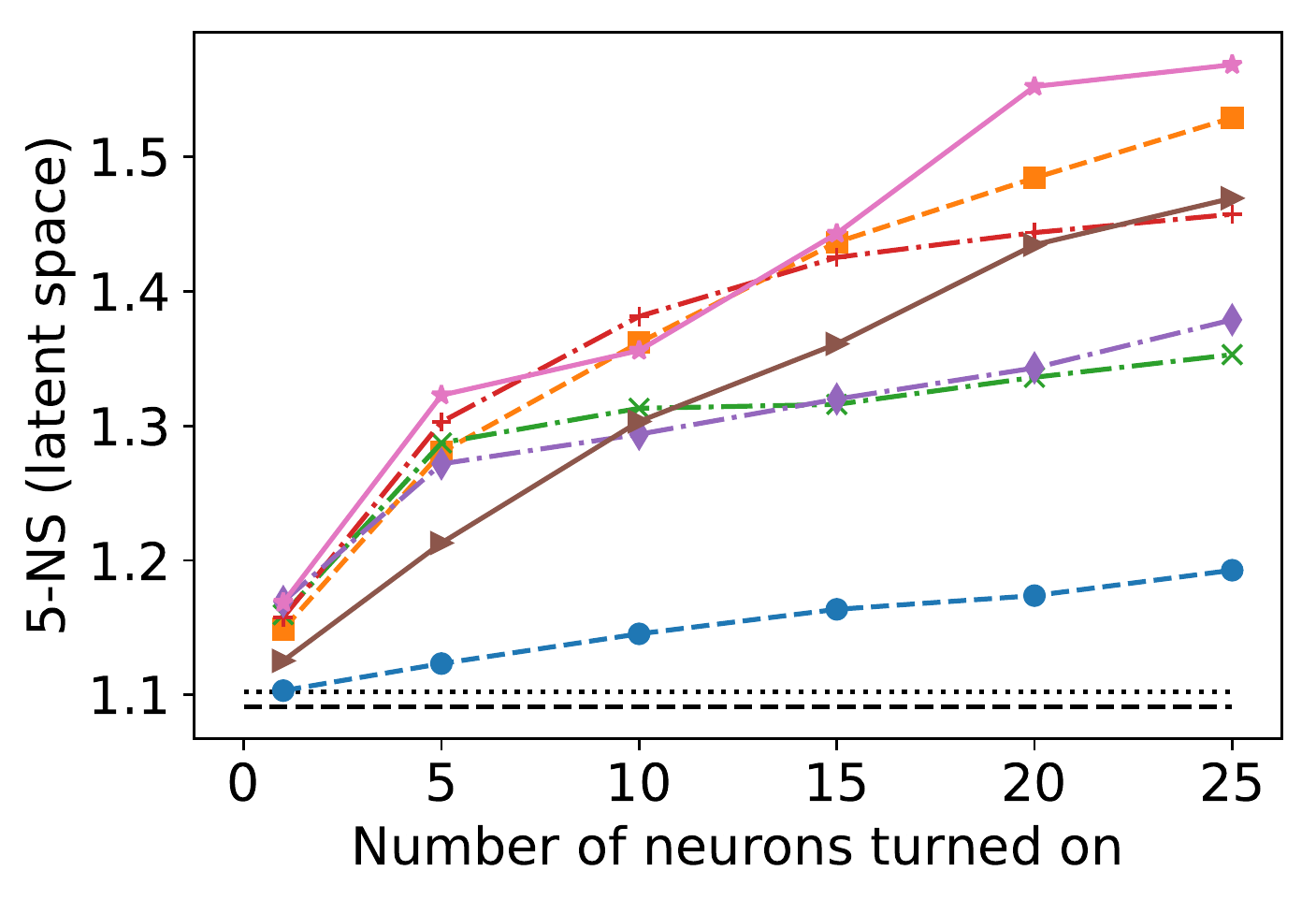}
        \caption{FMNIST}
    \end{subfigure}%
    \begin{subfigure}[b]{0.32\linewidth}
        \centering
        \includegraphics[width=\linewidth]{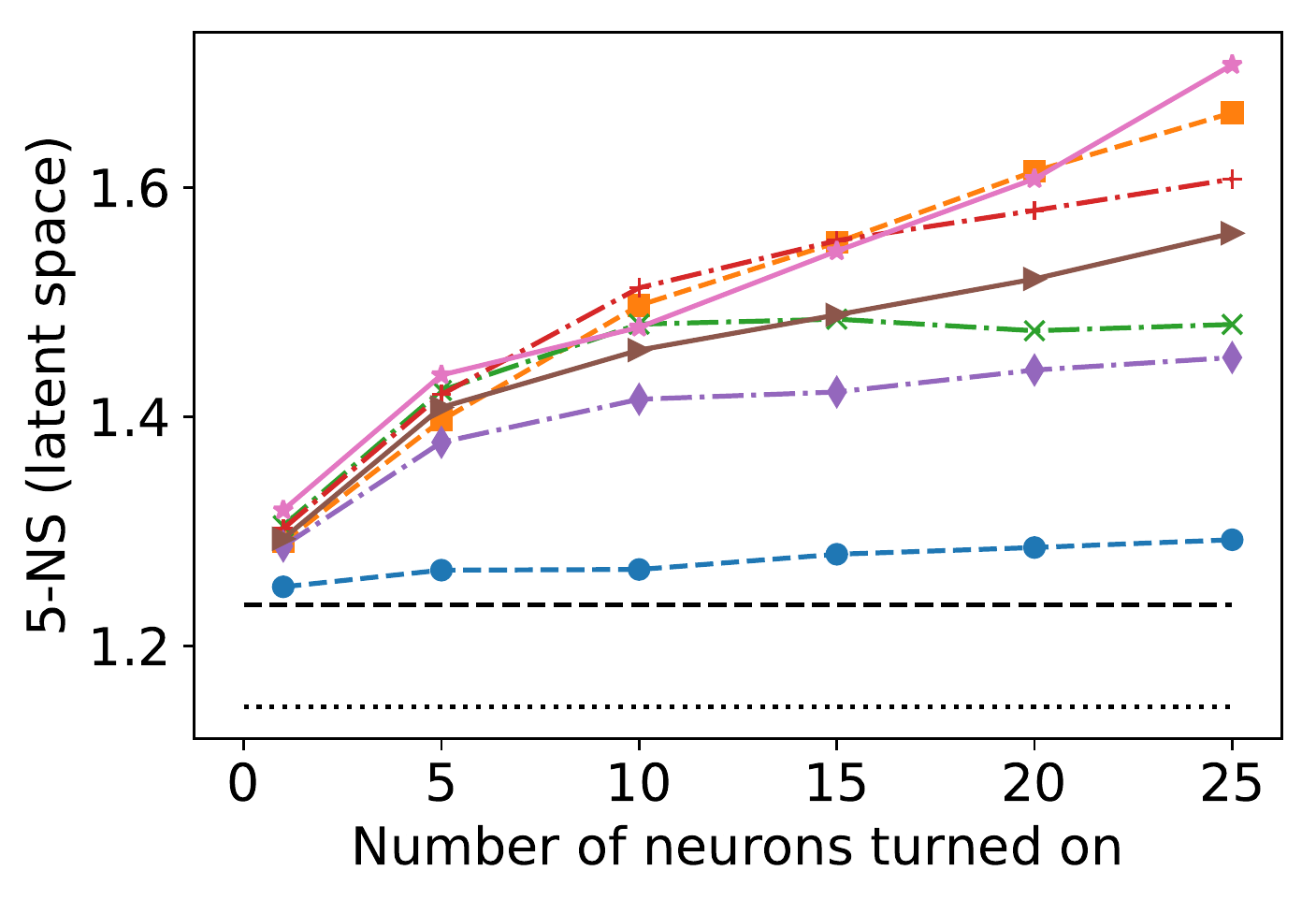}
        \caption{Combination}
    \end{subfigure}%
    \\
    \begin{subfigure}[b]{0.5\linewidth}
        \centering
        \includegraphics[width=\linewidth]{figs/legend.png}
        \caption{Legend}
    \end{subfigure}%
    \caption{Mean 5-NS (normalized $5^{th}$ nearest neighbor distance) in the latent space for each data set and approach, for increasing number of neurons turned on, from 1 to 25.}
   \label{fig:mean_norm5nn_latent_dist_per_nto}
\end{figure}

We analyzed the impact of the number of neurons being turned on on the various novelty-indicating metrics.  We varied the number of neurons turned on by each method from 1 to 25, generated 15000 samples with each method, and computed the metrics.  Metric results plots for each dataset and a few key metrics, for varying number of neurons turned on, are shown in Figures \ref{fig:mean_outlier_score_per_nto},  \ref{fig:mean_recon_dist_per_nto},  \ref{fig:mean_norm5nn_dist_per_nto}, and  \ref{fig:mean_norm5nn_latent_dist_per_nto}.

Complete metric results for varying numbers of neurons turned on are shown in Tables \ref{tab:results_per_neuron_mnist}, \ref{tab:results_per_neuron_fmnist}, and \ref{tab:results_per_neuron_combo} for MNIST, FMNIST, and MNIST+FMNIST combination, respectively.

We see that outlier/novelty metrics indicating outlier-ness tend to increase as more neurons are turned on across methods, sometimes up to a peak, after which when too many are turned on it may drop down and eventually level out for many methods.  This is because when more and more neurons get turned on, there becomes less variability in what is generated and eventually a point is reached where everything generated is equally as novel / garbage-like as it can get, and even further on a static image can be reached, e.g., if most or all neurons in a group are turned on every time.  

Furthermore we see that across metrics, the low-active (``loact'') method is among the top methods for different numbers of neurons turned on.  The entropy (``ent'') stands out for some as well, such as reconstruction distance, but in others is not among the top novel ones according to the metrics for increasing number of neurons turned on, for example, for in-domain score.  For in-domain score it starts off with a lower score (classified as more of an outlier) earlier and although it increases, it becomes surpassed by other methods.  This may be in part due to the fact that there are fewer neurons fitting the entropy category, so less variability in their effects.  Also, less extreme outlier classification is not necessarily a bad thing - as to achieve novelty a generated sample must have characteristics of an outlier, but if it becomes too extreme it also becomes useless / garbage - and so is not creative generation.

\begin{table*}[h]
    \centering
        \caption{Metric results for all approaches per number of neurons turned on (``num turn on'') for MNIST data.  }
    
    \scriptsize
    \begin{tabular}{lllllllllllllllll}
\toprule
 num turn on & approach &  oc score &  input 1NN mean &  input 5NN mean &  input 1NN norm mean &  input 5NN norm mean &  regen dists mean &  latent 1NN mean &  latent 5NN mean &  latent 1NN norm mean &  latent 5NN norm mean &  entropies mnist mean &  entropies fmnist mean &  fid mnist &  fid fmnist \\
\midrule
           0 &  2interp &    661.20 &           3.84 &           4.22 &                1.00 &                0.96 &              1.48 &            0.37 &            0.51 &                 1.12 &                 1.09 &                  0.07 &                   0.31 &     121.89 &      600.43 \\
           0 &  6interp &    888.69 &           4.43 &           4.73 &                1.05 &                0.98 &              1.61 &            0.32 &            0.46 &                 1.09 &                 1.08 &                  0.15 &                   0.37 &     305.54 &      658.93 \\
           0 &     test &       N/A &            N/A &            N/A &                 N/A &                 N/A &               N/A &             N/A &             N/A &                  N/A &                  N/A &                   N/A &                    N/A &      11.66 &      750.48 \\
           1 &   3clust &    489.60 &           4.15 &           4.53 &                1.11 &                1.06 &              2.66 &            0.43 &            0.59 &                 1.14 &                 1.12 &                  0.10 &                   0.30 &      78.17 &         N/A \\
           1 &    clust &    491.69 &           4.15 &           4.53 &                1.12 &                1.07 &              2.67 &            0.43 &            0.59 &                 1.14 &                 1.13 &                  0.10 &                   0.30 &      78.02 &         N/A \\
           1 &      cor &    513.12 &           4.17 &           4.53 &                1.12 &                1.07 &              2.68 &            0.42 &            0.58 &                 1.14 &                 1.12 &                  0.10 &                   0.30 &      86.05 &         N/A \\
           1 &      ent &    469.04 &           4.14 &           4.53 &                1.10 &                1.04 &              2.56 &            0.44 &            0.60 &                 1.14 &                 1.12 &                  0.08 &                   0.30 &      82.16 &         N/A \\
           1 &    loact &    496.31 &           4.39 &           4.75 &                1.18 &                1.12 &              3.08 &            0.43 &            0.60 &                 1.15 &                 1.13 &                  0.11 &                   0.29 &      76.06 &         N/A \\
           1 &    noise &    454.92 &           3.68 &           4.12 &                0.97 &                0.94 &              1.52 &            0.43 &            0.60 &                 1.12 &                 1.11 &                  0.05 &                   0.30 &      83.29 &         N/A \\
           1 &     rand &    486.12 &           4.13 &           4.50 &                1.11 &                1.06 &              2.60 &            0.43 &            0.60 &                 1.14 &                 1.12 &                  0.09 &                   0.30 &      76.21 &         N/A \\
           5 &   3clust &    562.12 &           4.96 &           5.25 &                1.44 &                1.38 &              4.39 &            0.46 &            0.62 &                 1.25 &                 1.21 &                  0.31 &                   0.27 &     122.81 &         N/A \\
           5 &    clust &    551.10 &           4.93 &           5.23 &                1.47 &                1.41 &              4.46 &            0.47 &            0.63 &                 1.27 &                 1.22 &                  0.30 &                   0.27 &     112.72 &         N/A \\
           5 &      cor &    562.56 &           4.94 &           5.22 &                1.50 &                1.44 &              4.66 &            0.48 &            0.63 &                 1.28 &                 1.23 &                  0.31 &                   0.26 &     126.57 &         N/A \\
           5 &      ent &    547.05 &           5.21 &           5.52 &                1.37 &                1.30 &              4.47 &            0.46 &            0.62 &                 1.24 &                 1.19 &                  0.24 &                   0.27 &     210.09 &         N/A \\
           5 &    loact &    525.94 &           5.22 &           5.54 &                1.46 &                1.38 &              4.87 &            0.49 &            0.65 &                 1.27 &                 1.23 &                  0.23 &                   0.21 &     154.13 &         N/A \\
           5 &    noise &    459.83 &           3.77 &           4.19 &                1.00 &                0.97 &              1.78 &            0.43 &            0.60 &                 1.12 &                 1.11 &                  0.06 &                   0.29 &      79.08 &         N/A \\
           5 &     rand &    551.11 &           4.89 &           5.18 &                1.41 &                1.35 &              4.21 &            0.46 &            0.61 &                 1.24 &                 1.19 &                  0.28 &                   0.27 &     111.61 &         N/A \\
          10 &   3clust &    551.05 &           5.10 &           5.36 &                1.63 &                1.57 &              4.85 &            0.52 &            0.67 &                 1.35 &                 1.29 &                  0.45 &                   0.23 &     175.77 &         N/A \\
          10 &    clust &    524.11 &           5.00 &           5.28 &                1.54 &                1.46 &              4.79 &            0.50 &            0.66 &                 1.33 &                 1.27 &                  0.29 &                   0.23 &     134.01 &         N/A \\
          10 &      cor &    481.46 &           4.91 &           5.21 &                1.50 &                1.42 &              4.72 &            0.50 &            0.66 &                 1.31 &                 1.25 &                  0.24 &                   0.22 &     119.32 &         N/A \\
          10 &      ent &    550.14 &           5.60 &           5.89 &                1.45 &                1.38 &              5.14 &            0.50 &            0.66 &                 1.32 &                 1.25 &                  0.35 &                   0.27 &     329.02 &         N/A \\
          10 &    loact &    483.64 &           5.38 &           5.70 &                1.53 &                1.46 &              5.15 &            0.53 &            0.68 &                 1.36 &                 1.28 &                  0.26 &                   0.18 &     238.20 &         N/A \\
          10 &    noise &    456.92 &           3.86 &           4.27 &                1.03 &                0.99 &              2.03 &            0.43 &            0.60 &                 1.12 &                 1.11 &                  0.07 &                   0.29 &      74.85 &         N/A \\
          10 &     rand &    540.34 &           5.06 &           5.32 &                1.60 &                1.53 &              4.71 &            0.51 &            0.66 &                 1.33 &                 1.28 &                  0.42 &                   0.23 &     167.52 &         N/A \\
          15 &   3clust &    500.42 &           5.09 &           5.35 &                1.70 &                1.64 &              4.95 &            0.56 &            0.72 &                 1.42 &                 1.34 &                  0.48 &                   0.21 &     209.25 &         N/A \\
          15 &    clust &    468.27 &           4.95 &           5.24 &                1.50 &                1.42 &              4.74 &            0.50 &            0.65 &                 1.32 &                 1.26 &                  0.21 &                   0.21 &     144.49 &         N/A \\
          15 &      cor &    421.53 &           4.81 &           5.14 &                1.45 &                1.38 &              4.55 &            0.49 &            0.65 &                 1.28 &                 1.22 &                  0.16 &                   0.21 &     107.27 &         N/A \\
          15 &      ent &    546.36 &           5.79 &           6.07 &                1.48 &                1.40 &              5.45 &            0.53 &            0.68 &                 1.36 &                 1.29 &                  0.40 &                   0.24 &     438.54 &         N/A \\
          15 &    loact &    438.95 &           5.41 &           5.72 &                1.59 &                1.52 &              5.21 &            0.57 &            0.73 &                 1.44 &                 1.34 &                  0.31 &                   0.17 &     315.60 &         N/A \\
          15 &    noise &    462.01 &           3.93 &           4.33 &                1.05 &                1.01 &              2.23 &            0.44 &            0.60 &                 1.14 &                 1.12 &                  0.08 &                   0.29 &      72.80 &         N/A \\
          15 &     rand &    501.47 &           5.01 &           5.26 &                1.72 &                1.66 &              4.81 &            0.56 &            0.72 &                 1.41 &                 1.34 &                  0.49 &                   0.21 &     217.08 &         N/A \\
          20 &   3clust &    451.76 &           5.06 &           5.32 &                1.74 &                1.68 &              4.95 &            0.60 &            0.76 &                 1.47 &                 1.38 &                  0.49 &                   0.20 &     232.18 &         N/A \\
          20 &    clust &    431.23 &           4.87 &           5.17 &                1.47 &                1.38 &              4.65 &            0.48 &            0.63 &                 1.30 &                 1.23 &                  0.14 &                   0.22 &     153.55 &         N/A \\
          20 &      cor &    375.39 &           4.74 &           5.09 &                1.42 &                1.34 &              4.42 &            0.48 &            0.65 &                 1.23 &                 1.20 &                  0.13 &                   0.20 &      98.08 &         N/A \\
          20 &      ent &    518.05 &           5.81 &           6.08 &                1.48 &                1.42 &              5.58 &            0.56 &            0.72 &                 1.40 &                 1.32 &                  0.50 &                   0.28 &     554.70 &         N/A \\
          20 &    loact &    408.31 &           5.41 &           5.71 &                1.67 &                1.58 &              5.28 &            0.61 &            0.77 &                 1.53 &                 1.38 &                  0.37 &                   0.19 &     373.35 &         N/A \\
          20 &    noise &    464.70 &           4.02 &           4.42 &                1.08 &                1.04 &              2.44 &            0.44 &            0.61 &                 1.13 &                 1.12 &                  0.09 &                   0.28 &      71.14 &         N/A \\
          20 &     rand &    446.04 &           4.91 &           5.15 &                1.82 &                1.75 &              4.80 &            0.61 &            0.77 &                 1.49 &                 1.39 &                  0.56 &                   0.20 &     260.45 &         N/A \\
          25 &   3clust &    414.89 &           5.04 &           5.30 &                1.76 &                1.70 &              4.94 &            0.62 &            0.79 &                 1.52 &                 1.42 &                  0.49 &                   0.19 &     247.96 &         N/A \\
          25 &    clust &    403.61 &           4.87 &           5.17 &                1.45 &                1.36 &              4.64 &            0.47 &            0.63 &                 1.27 &                 1.21 &                  0.11 &                   0.20 &     169.20 &         N/A \\
          25 &      cor &    346.43 &           4.67 &           5.02 &                1.40 &                1.32 &              4.33 &            0.49 &            0.65 &                 1.24 &                 1.19 &                  0.11 &                   0.19 &      96.57 &         N/A \\
          25 &      ent &    505.20 &           5.83 &           6.10 &                1.50 &                1.42 &              5.68 &            0.57 &            0.73 &                 1.43 &                 1.33 &                  0.52 &                   0.30 &     606.71 &         N/A \\
          25 &    loact &    391.35 &           5.38 &           5.66 &                1.74 &                1.61 &              5.34 &            0.64 &            0.80 &                 1.56 &                 1.41 &                  0.42 &                   0.21 &     427.99 &         N/A \\
          25 &    noise &    471.28 &           4.07 &           4.46 &                1.10 &                1.06 &              2.60 &            0.44 &            0.60 &                 1.14 &                 1.13 &                  0.10 &                   0.29 &      72.55 &         N/A \\
          25 &     rand &    392.39 &           4.80 &           5.05 &                1.88 &                1.80 &              4.76 &            0.65 &            0.82 &                 1.55 &                 1.43 &                  0.61 &                   0.20 &     297.59 &         N/A \\
\bottomrule
\end{tabular}
\label{tab:results_per_neuron_mnist}
\end{table*}

\begin{table*}[h]
    \centering
        \caption{Metric results for all approaches per number of neurons turned on (``num turn on'') for FMNIST data.  }
    
    \scriptsize
    \begin{tabular}{lllllllllllllllll}
\toprule
 num turn on & approach &  oc score &  input 1NN mean &  input 5NN mean &  input 1NN norm mean &  input 5NN norm mean &  regen dists mean &  latent 1NN mean &  latent 5NN mean &  latent 1NN norm mean &  latent 5NN norm mean &  entropies mnist mean &  entropies fmnist mean &  fid mnist &  fid fmnist \\
\midrule
           0 &  2interp &    817.61 &           2.79 &           3.11 &                0.97 &                0.92 &              1.21 &            0.21 &            0.31 &                 1.08 &                 1.09 &                  0.91 &                   0.27 &     651.27 &      249.92 \\
           0 &  6interp &   1012.52 &           2.68 &           2.96 &                0.99 &                0.94 &              1.36 &            0.16 &            0.24 &                 1.08 &                 1.10 &                  1.53 &                   0.48 &     675.79 &      442.53 \\
           0 &     test &       N/A &            N/A &            N/A &                 N/A &                 N/A &               N/A &             N/A &             N/A &                  N/A &                  N/A &                   N/A &                    N/A &     628.92 &       20.13 \\
           1 &   3clust &    671.54 &           3.92 &           4.23 &                1.28 &                1.20 &              3.17 &            0.26 &            0.38 &                 1.21 &                 1.16 &                  0.67 &                   0.25 &        N/A &      110.99 \\
           1 &    clust &    680.35 &           3.95 &           4.25 &                1.29 &                1.21 &              3.21 &            0.26 &            0.38 &                 1.19 &                 1.16 &                  0.68 &                   0.26 &        N/A &      112.75 \\
           1 &      cor &    704.78 &           3.88 &           4.21 &                1.21 &                1.17 &              3.11 &            0.26 &            0.38 &                 1.22 &                 1.17 &                  0.69 &                   0.28 &        N/A &      145.85 \\
           1 &      ent &    601.18 &           3.57 &           3.89 &                1.18 &                1.10 &              2.57 &            0.26 &            0.39 &                 1.15 &                 1.13 &                  0.67 &                   0.22 &        N/A &      155.63 \\
           1 &    loact &    678.56 &           4.35 &           4.62 &                1.43 &                1.33 &              3.77 &            0.26 &            0.38 &                 1.19 &                 1.17 &                  0.62 &                   0.23 &        N/A &      143.39 \\
           1 &    noise &    597.45 &           2.93 &           3.27 &                1.00 &                0.95 &              1.37 &            0.25 &            0.38 &                 1.11 &                 1.10 &                  0.70 &                   0.19 &        N/A &      173.29 \\
           1 &     rand &    657.37 &           3.80 &           4.11 &                1.24 &                1.17 &              2.97 &            0.26 &            0.38 &                 1.18 &                 1.15 &                  0.66 &                   0.24 &        N/A &      108.61 \\
           5 &   3clust &    785.92 &           5.10 &           5.39 &                1.54 &                1.50 &              5.07 &            0.29 &            0.39 &                 1.44 &                 1.30 &                  0.62 &                   0.36 &        N/A &      171.69 \\
           5 &    clust &    712.80 &           4.75 &           5.06 &                1.48 &                1.42 &              4.72 &            0.30 &            0.41 &                 1.40 &                 1.29 &                  0.58 &                   0.31 &        N/A &      143.34 \\
           5 &      cor &    722.28 &           4.52 &           4.87 &                1.34 &                1.32 &              4.43 &            0.29 &            0.40 &                 1.38 &                 1.27 &                  0.58 &                   0.32 &        N/A &      182.03 \\
           5 &      ent &    517.10 &           4.90 &           5.26 &                1.43 &                1.36 &              4.50 &            0.33 &            0.47 &                 1.26 &                 1.21 &                  0.32 &                   0.32 &        N/A &      261.75 \\
           5 &    loact &    574.25 &           5.45 &           5.78 &                1.64 &                1.59 &              5.49 &            0.34 &            0.46 &                 1.44 &                 1.32 &                  0.54 &                   0.16 &        N/A &      330.49 \\
           5 &    noise &    607.24 &           3.19 &           3.52 &                1.07 &                1.02 &              1.91 &            0.26 &            0.38 &                 1.14 &                 1.12 &                  0.21 &                   0.21 &        N/A &      133.91 \\
           5 &     rand &    774.91 &           5.03 &           5.31 &                1.55 &                1.49 &              4.93 &            0.28 &            0.39 &                 1.39 &                 1.28 &                  0.34 &                   0.34 &        N/A &      160.56 \\
          10 &   3clust &    785.61 &           5.31 &           5.61 &                1.57 &                1.56 &              5.51 &            0.31 &            0.41 &                 1.54 &                 1.38 &                  0.57 &                   0.35 &        N/A &      254.32 \\
          10 &    clust &    597.13 &           4.83 &           5.17 &                1.49 &                1.45 &              4.96 &            0.32 &            0.44 &                 1.40 &                 1.31 &                  0.49 &                   0.26 &        N/A &      179.84 \\
          10 &      cor &    620.14 &           4.76 &           5.14 &                1.40 &                1.39 &              4.87 &            0.32 &            0.43 &                 1.39 &                 1.29 &                  0.54 &                   0.30 &        N/A &      188.66 \\
          10 &      ent &    419.31 &           5.84 &           6.27 &                1.48 &                1.46 &              5.76 &            0.41 &            0.58 &                 1.34 &                 1.30 &                  0.63 &                   0.40 &        N/A &      376.76 \\
          10 &    loact &    467.08 &           5.66 &           6.00 &                1.66 &                1.63 &              5.70 &            0.36 &            0.49 &                 1.50 &                 1.36 &                  0.47 &                   0.15 &        N/A &      444.62 \\
          10 &    noise &    622.77 &           3.47 &           3.79 &                1.15 &                1.09 &              2.43 &            0.26 &            0.39 &                 1.17 &                 1.15 &                  0.68 &                   0.23 &        N/A &      112.54 \\
          10 &     rand &    804.73 &           5.32 &           5.61 &                1.57 &                1.57 &              5.45 &            0.30 &            0.40 &                 1.52 &                 1.36 &                  0.62 &                   0.37 &        N/A &      259.05 \\
          15 &   3clust &    763.77 &           5.34 &           5.65 &                1.58 &                1.59 &              5.64 &            0.33 &            0.43 &                 1.60 &                 1.43 &                  0.54 &                   0.33 &        N/A &      311.95 \\
          15 &    clust &    489.36 &           4.90 &           5.26 &                1.53 &                1.47 &              5.10 &            0.34 &            0.47 &                 1.40 &                 1.32 &                  0.45 &                   0.26 &        N/A &      210.21 \\
          15 &      cor &    510.42 &           4.90 &           5.30 &                1.44 &                1.44 &              5.13 &            0.35 &            0.47 &                 1.43 &                 1.32 &                  0.54 &                   0.30 &        N/A &      196.45 \\
          15 &      ent &    436.52 &           6.54 &           7.03 &                1.51 &                1.51 &              6.76 &            0.45 &            0.61 &                 1.42 &                 1.36 &                  0.56 &                   0.46 &        N/A &      514.98 \\
          15 &    loact &    385.89 &           5.35 &           5.68 &                1.62 &                1.66 &              5.31 &            0.41 &            0.55 &                 1.63 &                 1.44 &                  0.50 &                   0.13 &        N/A &      533.91 \\
          15 &    noise &    636.94 &           3.67 &           3.98 &                1.21 &                1.15 &              2.81 &            0.27 &            0.39 &                 1.19 &                 1.16 &                  0.68 &                   0.25 &        N/A &      105.50 \\
          15 &     rand &    799.23 &           5.30 &           5.59 &                1.57 &                1.58 &              5.49 &            0.32 &            0.41 &                 1.61 &                 1.44 &                  0.61 &                   0.36 &        N/A &      338.88 \\
          20 &   3clust &    734.59 &           5.35 &           5.68 &                1.58 &                1.60 &              5.71 &            0.34 &            0.44 &                 1.61 &                 1.44 &                  0.53 &                   0.31 &        N/A &      349.94 \\
          20 &    clust &    393.36 &           4.94 &           5.31 &                1.52 &                1.49 &              5.16 &            0.36 &            0.50 &                 1.40 &                 1.34 &                  0.44 &                   0.24 &        N/A &      240.97 \\
          20 &      cor &    407.01 &           5.00 &           5.41 &                1.46 &                1.46 &              5.25 &            0.37 &            0.51 &                 1.46 &                 1.34 &                  0.59 &                   0.30 &        N/A &      199.70 \\
          20 &      ent &    486.60 &           6.94 &           7.46 &                1.52 &                1.51 &              7.32 &            0.47 &            0.63 &                 1.45 &                 1.43 &                  0.51 &                   0.50 &        N/A &      663.10 \\
          20 &    loact &    298.18 &           4.86 &           5.18 &                1.53 &                1.62 &              4.82 &            0.46 &            0.61 &                 1.76 &                 1.55 &                  0.60 &                   0.14 &        N/A &      586.53 \\
          20 &    noise &    645.92 &           3.83 &           4.14 &                1.25 &                1.19 &              3.09 &            0.27 &            0.39 &                 1.23 &                 1.17 &                  0.66 &                   0.27 &        N/A &      102.85 \\
          20 &     rand &    780.84 &           5.24 &           5.52 &                1.56 &                1.59 &              5.45 &            0.33 &            0.43 &                 1.66 &                 1.48 &                  0.62 &                   0.36 &        N/A &      407.99 \\
          25 &   3clust &    708.28 &           5.35 &           5.69 &                1.57 &                1.62 &              5.75 &            0.35 &            0.45 &                 1.59 &                 1.46 &                  0.52 &                   0.30 &        N/A &      379.29 \\
          25 &    clust &    314.35 &           4.98 &           5.37 &                1.52 &                1.49 &              5.21 &            0.38 &            0.53 &                 1.42 &                 1.35 &                  0.45 &                   0.25 &        N/A &      268.52 \\
          25 &      cor &    319.62 &           5.09 &           5.51 &                1.47 &                1.48 &              5.37 &            0.41 &            0.55 &                 1.46 &                 1.38 &                  0.59 &                   0.31 &        N/A &      211.34 \\
          25 &      ent &    488.20 &           7.06 &           7.55 &                1.54 &                1.50 &              7.28 &            0.48 &            0.65 &                 1.40 &                 1.47 &                  0.50 &                   0.58 &        N/A &      781.26 \\
          25 &    loact &    250.80 &           4.35 &           4.65 &                1.50 &                1.56 &              4.24 &            0.47 &            0.62 &                 1.75 &                 1.57 &                  0.70 &                   0.21 &        N/A &      616.67 \\
          25 &    noise &    654.79 &           3.98 &           4.28 &                1.30 &                1.23 &              3.35 &            0.28 &            0.39 &                 1.24 &                 1.19 &                  0.67 &                   0.27 &        N/A &      103.42 \\
          25 &     rand &    764.71 &           5.16 &           5.45 &                1.55 &                1.60 &              5.39 &            0.34 &            0.44 &                 1.71 &                 1.53 &                  0.64 &                   0.34 &        N/A &      463.52 \\
\bottomrule
\end{tabular}
\label{tab:results_per_neuron_fmnist}
\end{table*}

\begin{table*}[h]
    \centering
    \caption{Metric results for all approaches per number of neurons turned on (``num turn on'') for FMNIST+MNIST combination data.  }
    \scriptsize
    \begin{tabular}{lllllllllllllllll}
\toprule
 num turn on & approach &  oc score &  input 1NN mean &  input 5NN mean &  input 1NN norm mean &  input 5NN norm mean &  regen dists mean &  latent 1NN mean &  latent 5NN mean &  latent 1NN norm mean &  latent 5NN norm mean &  entropies mnist mean &  entropies fmnist mean &  fid mnist &  fid fmnist \\
\midrule
           0 &  2interp &    745.21 &           3.93 &           4.28 &                1.13 &                1.10 &              1.62 &            1.04 &            1.24 &                 1.27 &                 1.24 &                  0.38 &                   0.28 &     245.49 &      260.23 \\
           0 &  6interp &    990.29 &           3.89 &           4.20 &                1.14 &                1.10 &              1.70 &            0.82 &            0.99 &                 1.18 &                 1.15 &                  0.76 &                   0.38 &     376.43 &      310.32 \\
           0 &     test &       N/A &            N/A &            N/A &                 N/A &                 N/A &               N/A &             N/A &             N/A &                  N/A &                  N/A &                   N/A &                    N/A &     217.92 &      238.62 \\
           1 &   3clust &    572.03 &           4.91 &           5.25 &                1.38 &                1.34 &              3.18 &            1.26 &            1.48 &                 1.34 &                 1.30 &                  0.30 &                   0.28 &     157.21 &      273.07 \\
           1 &    clust &    568.49 &           4.88 &           5.21 &                1.36 &                1.32 &              3.13 &            1.26 &            1.49 &                 1.34 &                 1.31 &                  0.30 &                   0.27 &     154.23 &      277.56 \\
           1 &      cor &    582.15 &           4.80 &           5.15 &                1.33 &                1.29 &              2.93 &            1.24 &            1.46 &                 1.33 &                 1.29 &                  0.32 &                   0.29 &     163.45 &      264.44 \\
           1 &      ent &    549.68 &           4.81 &           5.15 &                1.35 &                1.30 &              3.01 &            1.27 &            1.49 &                 1.35 &                 1.29 &                  0.30 &                   0.28 &     173.47 &      266.82 \\
           1 &    loact &    583.91 &           5.32 &           5.62 &                1.50 &                1.45 &              3.92 &            1.27 &            1.49 &                 1.38 &                 1.32 &                  0.29 &                   0.27 &     153.69 &      319.03 \\
           1 &    noise &    531.79 &           4.17 &           4.56 &                1.15 &                1.13 &              1.68 &            1.24 &            1.47 &                 1.28 &                 1.25 &                  0.32 &                   0.26 &     194.92 &      282.60 \\
           1 &     rand &    558.37 &           4.85 &           5.19 &                1.36 &                1.31 &              3.09 &            1.26 &            1.49 &                 1.35 &                 1.29 &                  0.30 &                   0.27 &     153.26 &      273.92 \\
           5 &   3clust &    588.93 &           5.66 &           5.93 &                1.64 &                1.61 &              4.71 &            1.34 &            1.55 &                 1.48 &                 1.42 &                  0.37 &                   0.26 &     132.72 &      368.98 \\
           5 &    clust &    563.03 &           5.56 &           5.83 &                1.61 &                1.57 &              4.53 &            1.37 &            1.58 &                 1.49 &                 1.42 &                  0.33 &                   0.26 &     125.76 &      352.20 \\
           5 &      cor &    528.91 &           5.33 &           5.65 &                1.49 &                1.45 &              4.07 &            1.36 &            1.58 &                 1.44 &                 1.38 &                  0.32 &                   0.30 &     135.00 &      313.69 \\
           5 &      ent &    512.17 &           5.57 &           5.85 &                1.66 &                1.60 &              4.52 &            1.39 &            1.60 &                 1.48 &                 1.41 &                  0.27 &                   0.27 &     193.68 &      344.90 \\
           5 &    loact &    488.66 &           5.65 &           5.96 &                1.62 &                1.56 &              4.91 &            1.43 &            1.66 &                 1.50 &                 1.44 &                  0.26 &                   0.16 &     189.25 &      529.79 \\
           5 &    noise &    537.48 &           4.29 &           4.67 &                1.19 &                1.17 &              1.97 &            1.25 &            1.47 &                 1.30 &                 1.27 &                  0.26 &                   0.26 &     178.34 &      278.48 \\
           5 &     rand &    583.58 &           5.66 &           5.93 &                1.63 &                1.59 &              4.68 &            1.33 &            1.54 &                 1.46 &                 1.40 &                  0.26 &                   0.26 &     128.40 &      357.65 \\
          10 &   3clust &    502.66 &           5.56 &           5.81 &                1.71 &                1.69 &              4.81 &            1.47 &            1.68 &                 1.59 &                 1.51 &                  0.43 &                   0.22 &     146.86 &      445.96 \\
          10 &    clust &    447.39 &           5.37 &           5.65 &                1.65 &                1.62 &              4.56 &            1.50 &            1.73 &                 1.54 &                 1.48 &                  0.36 &                   0.24 &     124.72 &      423.00 \\
          10 &      cor &    417.83 &           5.39 &           5.72 &                1.55 &                1.50 &              4.32 &            1.48 &            1.71 &                 1.48 &                 1.42 &                  0.34 &                   0.28 &     131.99 &      347.27 \\
          10 &      ent &    464.15 &           5.55 &           5.81 &                1.79 &                1.72 &              4.82 &            1.45 &            1.67 &                 1.55 &                 1.46 &                  0.34 &                   0.24 &     213.11 &      422.31 \\
          10 &    loact &    385.08 &           5.65 &           5.95 &                1.67 &                1.59 &              4.94 &            1.57 &            1.80 &                 1.57 &                 1.48 &                  0.27 &                   0.15 &     230.01 &      645.99 \\
          10 &    noise &    537.70 &           4.41 &           4.78 &                1.23 &                1.20 &              2.26 &            1.25 &            1.48 &                 1.30 &                 1.27 &                  0.31 &                   0.26 &     165.17 &      278.84 \\
          10 &     rand &    505.89 &           5.59 &           5.84 &                1.72 &                1.69 &              4.83 &            1.47 &            1.68 &                 1.56 &                 1.50 &                  0.45 &                   0.22 &     141.63 &      443.59 \\
          15 &   3clust &    398.25 &           5.33 &           5.58 &                1.74 &                1.73 &              4.65 &            1.59 &            1.81 &                 1.62 &                 1.55 &                  0.47 &                   0.20 &     167.64 &      499.03 \\
          15 &    clust &    357.17 &           5.21 &           5.49 &                1.65 &                1.60 &              4.42 &            1.58 &            1.81 &                 1.54 &                 1.49 &                  0.38 &                   0.22 &     133.81 &      473.77 \\
          15 &      cor &    341.22 &           5.47 &           5.81 &                1.56 &                1.51 &              4.48 &            1.54 &            1.79 &                 1.48 &                 1.42 &                  0.37 &                   0.26 &     141.25 &      361.79 \\
          15 &      ent &    388.43 &           5.38 &           5.63 &                1.81 &                1.74 &              4.72 &            1.53 &            1.76 &                 1.58 &                 1.49 &                  0.34 &                   0.24 &     234.36 &      468.85 \\
          15 &    loact &    280.54 &           5.56 &           5.85 &                1.75 &                1.68 &              4.91 &            1.70 &            1.95 &                 1.65 &                 1.54 &                  0.32 &                   0.15 &     279.24 &      707.14 \\
          15 &    noise &    547.22 &           4.52 &           4.88 &                1.27 &                1.24 &              2.54 &            1.26 &            1.48 &                 1.32 &                 1.28 &                  0.33 &                   0.27 &     160.99 &      274.81 \\
          15 &     rand &    409.38 &           5.36 &           5.60 &                1.74 &                1.75 &              4.66 &            1.58 &            1.80 &                 1.63 &                 1.55 &                  0.49 &                   0.21 &     169.47 &      497.30 \\
          20 &   3clust &    307.47 &           5.19 &           5.44 &                1.74 &                1.74 &              4.52 &            1.68 &            1.92 &                 1.66 &                 1.58 &                  0.50 &                   0.19 &     188.69 &      533.25 \\
          20 &    clust &    288.63 &           5.09 &           5.39 &                1.62 &                1.57 &              4.29 &            1.63 &            1.88 &                 1.56 &                 1.48 &                  0.44 &                   0.21 &     149.59 &      517.82 \\
          20 &      cor &    278.79 &           5.50 &           5.85 &                1.59 &                1.54 &              4.56 &            1.60 &            1.85 &                 1.53 &                 1.44 &                  0.39 &                   0.25 &     158.18 &      371.81 \\
          20 &      ent &    326.44 &           5.22 &           5.46 &                1.84 &                1.77 &              4.60 &            1.58 &            1.82 &                 1.60 &                 1.52 &                  0.34 &                   0.23 &     254.74 &      504.00 \\
          20 &    loact &    188.55 &           5.30 &           5.60 &                1.79 &                1.74 &              4.71 &            1.82 &            2.07 &                 1.75 &                 1.61 &                  0.38 &                   0.16 &     324.56 &      731.92 \\
          20 &    noise &    541.79 &           4.60 &           4.95 &                1.30 &                1.27 &              2.73 &            1.26 &            1.49 &                 1.33 &                 1.29 &                  0.33 &                   0.27 &     151.59 &      280.86 \\
          20 &     rand &    305.25 &           5.09 &           5.34 &                1.74 &                1.76 &              4.42 &            1.70 &            1.93 &                 1.69 &                 1.61 &                  0.54 &                   0.19 &     202.96 &      536.41 \\
          25 &   3clust &    235.30 &           5.05 &           5.31 &                1.75 &                1.75 &              4.40 &            1.76 &            2.00 &                 1.69 &                 1.61 &                  0.51 &                   0.18 &     207.16 &      554.97 \\
          25 &    clust &    244.32 &           5.04 &           5.35 &                1.61 &                1.55 &              4.24 &            1.68 &            1.93 &                 1.59 &                 1.48 &                  0.46 &                   0.19 &     167.04 &      552.28 \\
          25 &      cor &    227.55 &           5.49 &           5.84 &                1.60 &                1.55 &              4.59 &            1.64 &            1.89 &                 1.55 &                 1.45 &                  0.38 &                   0.24 &     167.58 &      383.76 \\
          25 &      ent &    269.26 &           5.07 &           5.31 &                1.84 &                1.78 &              4.48 &            1.65 &            1.90 &                 1.64 &                 1.56 &                  0.38 &                   0.24 &     273.58 &      534.75 \\
          25 &    loact &     93.76 &           5.00 &           5.31 &                1.77 &                1.77 &              4.45 &            1.97 &            2.22 &                 1.87 &                 1.71 &                  0.37 &                   0.16 &     370.79 &      739.24 \\
          25 &    noise &    537.78 &           4.66 &           5.00 &                1.32 &                1.29 &              2.86 &            1.28 &            1.51 &                 1.34 &                 1.29 &                  0.32 &                   0.26 &     143.24 &      289.18 \\
          25 &     rand &    210.75 &           4.84 &           5.09 &                1.72 &                1.77 &              4.20 &            1.80 &            2.04 &                 1.74 &                 1.67 &                  0.59 &                   0.18 &     239.57 &      562.57 \\
\bottomrule
\end{tabular}
    
    \label{tab:results_per_neuron_combo}
\end{table*}

\subsection{More details and analysis of interpolation method}

\begin{figure*}[t]
\centering
\begin{center}
\centering
\centerline{\includegraphics[width=1.9\columnwidth]{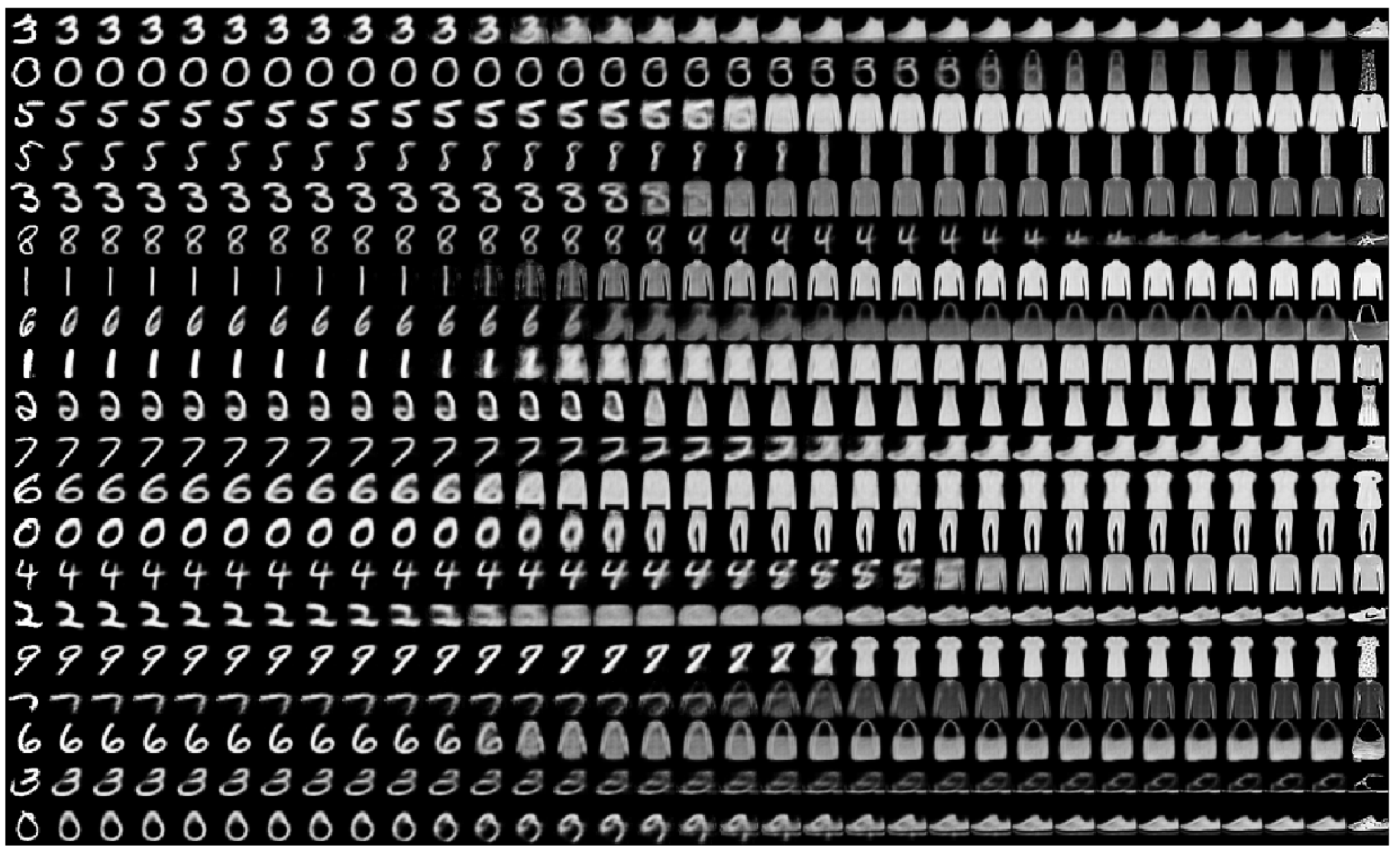}}
\caption{Illustration of images generated by interpolation in the latent space between two different classes of images, for the combined MNIST+FMNIST dataset.  Left is the MNIST image, right the FMNIST image, in between are generated image along the interpolation path between the two in the latent (z) space}
\label{fig:interp}
\end{center}
\end{figure*}

For the interpolation methods we compared with, we performed linear interpolation in $z$ space between a number (2 and 6) of random points that belong to different classes and evaluated the images at the mid-point of interpolation.  As those images show the opposite trend in terms of surrogate metrics (Figures 8-10 and Tables 4-6, 2interp and 6interp), they  were not considered for human evaluation.  Sample images are shown in Figure \ref{fig:interp}.   We see that usually interpolation does not yield anything much different from regular generation, except for around the midpoints, in which case it often does not seem to introduce new components in a meaningful way.  We also noted that there is even less difference from regular generation, even in the midpoints, for the MNIST and FMNIST datasets alone - likely due to the limited variability.  This ie expected based on the nature of VAEs as de-noising auto-encoders / the objective of VAEs to encode nearby points to the same final generated image - most of the space nearby what generates a recognizable will be pulled to generate the same image.  We also tried decoding from random points along interpolation paths, which we yielded worse results.  

\subsection{Results from previous user study}
Here we include results from a prior, more problematic study - in which too many options were given to the user, and appropriate categories were not chosen.  Hence we revised the human study and show the results in the main text.  Nevertheless, this earlier study with different participants corroborates out conclusions.

\textbf{Human annotation.} Figures \ref{Creativity}E and \ref{Human}A present the analysis of human annotations of the generated samples by 10 annotators. For comparison, samples that were annotated as ``not creative'' and ``garbage'' are also shown (Fig. \ref{Human}B-C). Visually, the creative samples generated from the latent space of MNIST digits do not appear digit-like anymore, but look more like symbols. In contrast, those generated from FMNIST still resemble fashion objects; however, novel objects such as a jumpsuit, sock, shoe with two heels, or front-slit kurta were found. Comparison with ``garbage'' images confirms the known association of both novelty and value with the human perception of creativity. Interestingly, the VAE trained on the combined dataset outputs creative images that are distinct from the MNIST-only and FMNIST-only cases, suggesting the dependency of creative exploration on the ``richness'' or diversity of the underlying concept space. It must be noted that the present perception of “creativity” is mostly aesthetical, so experiments on more complex ``artsy" datasets are needed and will be performed in future. We estimated Fleiss' $\kappa$ for human annotations, which are -0.47, -0.46, and -0.38 for MNIST, FMNIST, and combined, respectively. This result is interpreted as low agreements among users, which likely represent the subjective nature of creativity/novelty evaluation.

\begin{table*}[hbt!]
\small
\setlength\tabcolsep{4.5pt}
\centering
\caption{Ranking  for different decoding schemes according to human perception in prior human study: L1 (Creative), L2 (Garbage), L3 (Not Creative), L4 (Not sure). Values reported are the normalized fraction of annotated instances (with majority vote)) within each decoding scheme; (filter) indicates the filtered version.}

\begin{tabular}{l|cccc|cccc|cccc}
\hline 
  & \multicolumn{4}{c|}{MNIST } & \multicolumn{4}{c|}{FMNIST } & \multicolumn{4}{c}{MNIST+FMNIST }\tabularnewline
\hline 
  & L1  & L2  & L3   & L4  & L1  & L2  & L3  & L4  & L1  & L2  & L3   & L4 \tabularnewline
\hline
Non-specific & 0.29 & 0.35 & 0.13 & 0.24 & 0.21 & 0.38 & 0.31 & 0.102 & 0.27 & 0.52 & 0.15 & 0.06 \tabularnewline 
Non-specific(filter) & 0.33 & 0.39 & 0.05 & 0.23 & 0.25 & 0.41 & 0.25 & 0.085 & 0.27 & 0.50 & 0.13 & 0.09 \tabularnewline 
Low-active & 0.34 & 0.36 & 0.12 & 0.18 & 0.18 & 0.53 & 0.21 & 0.077 & 0.37 & 0.43 & 0.14 & 0.06 \tabularnewline 
Low-active(filter) & 0.47 & 0.41 & 0.03 & 0.09 & 0.29 & 0.42 & 0.21 & 0.090 & 0.36 & 0.44 & 0.10 & 0.10 \tabularnewline 
Noisy & 0.08 & 0.08 & 0.80 & 0.05 & 0.11 & 0.16 & 0.67 & 0.062 & 0.24 & 0.15 & 0.49 & 0.11 \tabularnewline 
Random & 0.23 & 0.32 & 0.25 & 0.21 & 0.11 & 0.59 & 0.25 & 0.049 & 0.14 & 0.60 & 0.14 & 0.13 \tabularnewline 
Regular & 0.13 & 0.03 & 0.83 & 0.01 & 0.07 & 0.07 & 0.85 & 0.013 & 0.29 & 0.11 & 0.56 & 0.04 \tabularnewline

\hline 
\end{tabular}

\label{old_ranking}
\end{table*}

\textbf{Comparison with baseline methods.} While comparing  different decoding schemes in terms of creative capacity, as estimated by the normalized fraction of creative samples with low subject variability (shown in Table \ref{old_ranking}, L1 columns), the low-active method  constantly outperformed the baseline methods (random flipping of ``off'' neurons, noisy decoding) and regular decoding  in terms of creative generation. And the gain clearly depends on the dataset. Low-active decoding from MNIST (or combined) data more frequently results in creative samples. In contrast,  creative generation from FMNIST space alone seems more challenging, possibly due to the value assessment of generated  objects in the context of fashion, resulting in higher fractions of ``garbage'' annotation than ``creative''.   
This trend is again reflected in the generations resulting from creative decoding from the combined MNIST+FMNIST space, as shown in Figure \ref{transform} (here we see shifting and deletion of original features often occurred as a result of creative decoding; new, unusual features were also added). The enhanced complexity in the space of  combined data helps creative generation even in baseline methods; however, low-active decoding wins again.  In contrast, while non-specific method performs well in MNIST and FMINST, it performs similar to regular decoding for the combimed dataset and also yields images that are categorized as garbage.  This is likely due to the fact that this method essentially tries to generate images in a class non-specific manner. And with increased complexity of the dataset, there is a higher risk of annotators not recognizing the generated images by this method as a valid object (not belonging to a known class) and categorizing is as not ``useful'', \textit{i.e.}, garbage.


Noisy decoding performs similar to regular decoding, while random  activation of additional ``off'' neurons has a high tendency to output either non-creative or garbage . 

Table \ref{old_ranking} also shows that novelty metrics can aid in selecting high fidelity creative images for MNIST and FMNIST.   

\begin{table*}[hbt!]
\small
\setlength\tabcolsep{4pt}
\centering
\caption{Comparison between human judgment  and novelty metrics: Novelty score  $NS(z)$ (considering top 5 nearest neighbors),  reconstruction distance $D_{r}$, in-domain MNIST classifier entropy $ICE_{M}$,  in-domain FMNIST classifier entropy $ICE_{F}$, in-domain score ($IS$) obtained using  one-class SVM classifier. L1 (Creative), L2 (Garbage), L3 (Not Creative), L4 (Not sure).}

\small
\scalebox{1.0}[1.0] {
\begin{tabular}{l|cccc|cccc|ccccc}
\hline 
 & \multicolumn{4}{c|}{MNIST} & \multicolumn{4}{c|}{FMNIST} & \multicolumn{5}{c}{MNIST+FMNIST}\tabularnewline
\hline 
  & $NS_{z}$  & $D_{r}$  & $ICE_M$  & $IS$  & $NS_{z}$  & $D_{r}$ & $ICE_F$ & $IS$  & $NS_{z}$ & $D_{r}$  & $ICE_M$ & $ICE_F$  & $IS$\tabularnewline
\hline 
L1 & 1.28 & 5.53 & 0.17 & 425.02 & 1.36 & 5.61 & 0.19 & 253.96 & 1.37 & 4.81 & 0.18 & 0.19 & 318.10 \tabularnewline 
L2 & 1.29 & 5.42 & 0.27 & 434.63 & 1.39 & 5.59 & 0.25 & 540.98 & 1.48 & 5.40 & 0.29 & 0.22 & 342.82 \tabularnewline 
L3 & 1.15 & 3.04 & 0.13 & 469.38 & 1.26 & 4.08 & 0.22 & 497.16 & 1.29 & 3.90 & 0.24 & 0.20 & 422.07 \tabularnewline 
L4 & 1.23 & 5.12 & 0.29 & 508.36 & 1.32 & 5.41 & 0.19 & 405.97 & 1.45 & 4.76 & 0.31 & 0.28 & 359.33 \tabularnewline 

\hline 

\end{tabular}
}

\label{old_humanvsml}
\end{table*}

\textbf{Relation between human judgment and novelty metrics.} Next, we report the values of novelty  metrics and their relation with the human judgment of creativity in Table \ref{old_humanvsml}. The standard deviations  (not shown) are small for $Dr$ and $kNSz$ (\textit{e.g.}, std is 20\% of the average $D_r$), while ICE and IS show slightly higher variability.  Nevertheless, the overall trend of ML metrics between creative and not-creative images remains consistent across all datasets.
Table \ref{old_humanvsml} reveals that a combination of  lower (or equal) in-domain classifier entropy, lower in-domain score, and higher $D_r$ and $kNS_z$ with respect to the not-creative samples are key characteristics of the creative images, on average. These results are consistent with  earlier findings \cite{cherti2017out},  demonstrating that out-of-class metrics capture well the creative capacity of generative models. On the other hand, ``not sure'' and garbage images are mostly characterized by  higher $kNS_z$ and slightly higher $ICE$.    

We further trained a ``creativity'' classifier using the surrogate metrics (Table 4) as features, and found a combination of metrics provides superior predictability of creativity over any single metric.  For example, a trained L1-regularized logistic regression classifier  for predicting consensus "garbage", "creative" or "not creative" yields a 10-fold CV mean accuracy of 71.0\% whereas the best single-metric accuracy was 60.7\% on  FMNIST evaluations.

We  report  the full analysis of novelty metrics for several generation methods (including cluster-based and correlation-based flipping as well as linear interpolation in latent space between different classes), as a function of varying numbers of flipping neurons in the Supplementary Materials (Tables 4-6).  This systematic analysis reveals a general increase of  out-of-domain nature and novelty score as number of flipped neurons increases. Nevertheless,  this effect seems to level off at higher numbers ($\geq$ 20) of neurons  - possibly indicating that the generated samples  saturate in terms of distortion beyond a certain level. In accordance with human evaluation, the trend of surrogate metrics indicative of novelty also ranks the neuro-inspired low-active method (and non-specific variant) as the more efficient approach in terms of novelty generation capacity,  particularly when the number of flipped neurons is minimal. Figures 9-12 and Tables 4-6 further demonstrate that activating multiple ``off" neuronal clusters together (3clust results) is effective as well, emphasizing need for atypical neuronal coordination for novelty generation. The results from linear interpolation in the $z$ space showed  opposite trend in terms of surrogate metrics (Figures 9-11 and Tables 4-6, 2interp and 6interp) and visual inspection confirmed that the resulting samples were not novel (Figure 13). 
It should also be mentioned, that the special effect of ``creative'' decoding cannot be replicated by simply flipping random ``off'' neurons (Table \ref{ranking} and novelty metrics analysis section in Supplementary Materials) - the neuro-inspired selection and flipping of low-active neurons is what promotes creativity in generations. 

\raggedbottom

\end{document}